\documentclass{article}

\usepackage{microtype}
\usepackage{graphicx}
\usepackage{subcaption}
\usepackage{booktabs} 
\usepackage{siunitx}
\usepackage{bm}
\usepackage{amsmath}
\usepackage{dsfont}

\usepackage{hyperref}

\usepackage[accepted]{icml2019}

\begin{document}

\twocolumn[
\icmltitle{Decoding Molecular Graph Embeddings with Reinforcement Learning}



\icmlsetsymbol{equal}{*}

\begin{icmlauthorlist}
\icmlauthor{Steven Kearnes}{google}
\icmlauthor{Li Li}{google}
\icmlauthor{Patrick Riley}{google}
\end{icmlauthorlist}

\icmlaffiliation{google}{Google Research, Mountain View, California, USA}

\icmlcorrespondingauthor{Steven Kearnes}{kearnes@google.com}

\icmlkeywords{Machine Learning, ICML}

\vskip 0.3in
]



\printAffiliationsAndNotice{}  

\begin{abstract}

We present RL-VAE, a graph-to-graph variational autoencoder that uses reinforcement learning to decode molecular graphs from latent embeddings. Methods have been described previously for graph-to-graph autoencoding, but these approaches require sophisticated decoders that increase the complexity of training and evaluation (such as requiring parallel encoders and decoders or non-trivial graph matching). Here, we repurpose a simple graph generator to enable efficient decoding and generation of molecular graphs.

\end{abstract}

\section{Introduction}


There are two prevailing approaches for \emph{in silico} molecular property optimization---molecular autoencoders and generative models---and there is a very real division in the field depending on whether the optimization happens in the latent space of an autoencoder or by conditioning a generative process.
Both types of models have advantages and disadvantages. Autoencoders have a latent space that naturally lends itself to continuous optimization; this was elegantly demonstrated by \citet{Gomez-Bombarelli2018-yy}. On the other hand, these models struggle to decode valid molecular graphs due to their discrete nature and the constraints of chemistry. Early work focused on character-level autoencoders trained on SMILES strings \cite{Weininger1988-kf}---for example, caffeine is represented as \texttt{CN1C=NC2=C1C(=O)N(C(=O)N2C)C}---but the syntactic complexity and long-range dependencies of SMILES make decoding particularly difficult. Subsequent models incorporated grammatical and syntactic constraints to avoid generating invalid strings \cite{Dai2018-ub, Kusner2017-dr}.

Attempts to decode graphs directly (\emph{i.e.}, without a SMILES intermediate) have led to useful but complicated models. For example, \citet{Jin2018-rz} describe a model that requires reducing each molecule to a tree structure and combining information from two parallel decoders to expand a predicted tree to a valid graph, and \citet{Simonovsky2018-xl} predict entire graphs in one step but require complicated graph-matching and heuristics to avoid disconnected or invalid results.

Recent work in generative models has sidestepped the need for differentiable graph-isomorphic loss functions that makes graph decoders awkward, instead borrowing methods from reinforcement learning to iteratively construct chemically valid graphs according to a learned value function. Examples include \citet{You2018-ym, Li2018-oi, Li2018-zm, Zhou2018-yc}, which all formulate graph generation as a Markov decision process (MDP) and learn policies for generating molecules that match specific criteria. While effective for generation and optimization, these methods sacrifice the notion of a continuous latent space with potentially useful structure \cite{Mikolov2013-sr}.

Here we describe a model, RL-VAE (Reinforcement Learning Variational Autoencoder), which combines advantages from both types of models: specifically, the model has a latent space that enables continuous optimization while using an iterative MDP-based graph decoder that guarantees chemically valid output. To the best of our knowledge, this is the first example of a variational autoencoder
that uses a reinforcement learning agent as the decoder.

\section{Methods}

\begin{figure}[tb]
\begin{center}
\begin{subfigure}[b]{0.40\linewidth}
\centerline{\includegraphics[width=\linewidth]{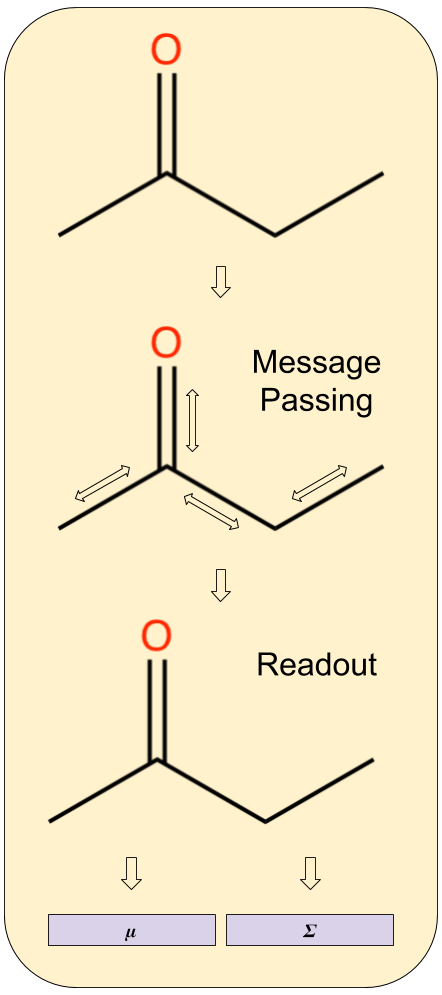}}
\caption{Encoder.}
\end{subfigure}
\begin{subfigure}[b]{0.40\linewidth}
\centerline{\includegraphics[width=\linewidth]{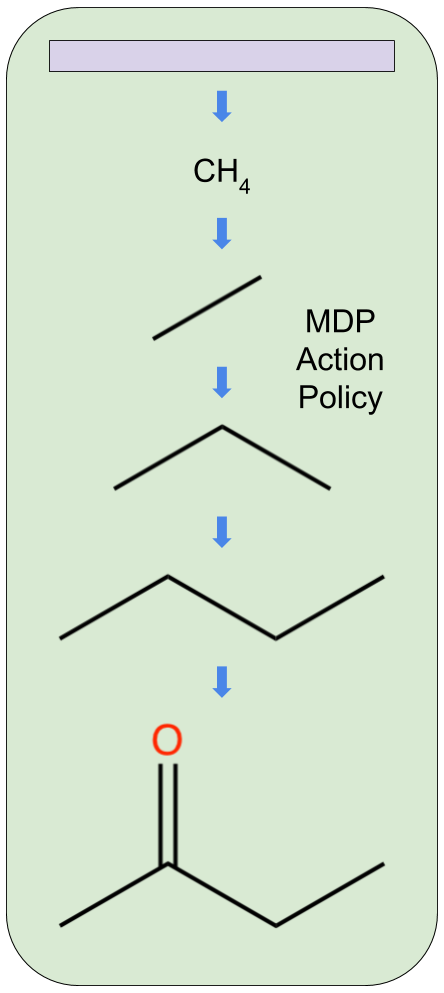}}
\caption{Decoder.}
\end{subfigure}
\caption{Graphical description of the RL-VAE architecture. (a)~The encoder converts a molecular graph into an embedding distribution parameterized by $\mu$ and $\Sigma$ using a message passing neural network (MPNN). (b)~The decoder samples a vector from the embedding distribution and decodes it into a molecular graph using a learned value function for a molecule-specific Markov decision process (MDP).}
\label{fig:rl-vae}
\end{center}
\vskip -0.2in
\end{figure}

The overall architecture of the RL-VAE is depicted in \figurename~\ref{fig:rl-vae}. The following subsections detail the model and our training procedures.



\subsection{Graph Encoder}
\label{encoder}

Input molecular graphs were represented with node- and edge-level features: one-hot encodings of atom and bond types, respectively. The set of allowed atoms was \{H, C, N, O, F\}; the set of allowed bonds was \{single, double, triple, aromatic\}. No other input features were used. The node and edge features were linearly mapped (without any bias terms) to a learned latent space with dimension 128 prior to being fed into the encoder.

Molecular graphs were encoded using a message passing neural network (MPNN) \cite{Gilmer2017-yn}. In particular, we followed the description from \citet{Li2018-zm}, including a gating network that determines the relative contribution of each node to the graph-level representation.

The MPNN used two graph convolutional layers; the layers did not apply any learned transformations to the messages, but a gated recurrent unit (GRU) \cite{Cho2014-ps} with learned parameters was used to update node states with the aggregated messages from neighboring nodes. The final node features were mapped to 256-dimensional vectors with a learned linear transform, gated, and summed to produce the graph-level representation (this process is referred to as a ``readout'' transformation). Two readout transformations with separate weights were used to predict the mean $\mu$ and (log) standard deviation $\Sigma$ used for variational sampling of the latent space \cite{Doersch2016-md}. The final embedding space thus had \num{256} dimensions; molecules were embedded as distributions---parameterized by $\mu$ and $\Sigma$---from which individual molecular embeddings were sampled.

\subsection{Graph Decoder}
\label{sec:decoder}

Molecular embeddings were decoded using a variant of MolDQN, a graph generator that uses a Markov decision process (MDP) to construct graphs sequentially \cite{Zhou2018-yc}. The model was trained with Double $Q$-learning \cite{Van_Hasselt2015-my} to approximate $V(s)$, the state value function.

The MDP was modified slightly from \citet{Zhou2018-yc}. Specifically: (1) bonds could not be removed or promoted (\emph{i.e.}, the model was not allowed to backtrack or update earlier steps); (2) when two single-atom fragments were created by removing a bond, both fragments were added to the action set (the original implementation added only the first fragment in a sorted list of SMILES); and (3) our implementation did not add triple bonds between existing atoms in the graph (to avoid forming rings containing triple bonds). Notably, this MDP guarantees that all generated molecules are chemically valid.

To condition the generative process to reconstruct specific molecules, an embedding of the target molecule was concatenated to the embedding of the current state as input to the value function. Additionally, the value function received information about the number of steps remaining in the episode, as well as whether the proposed action lead to a terminal state (\emph{i.e.}, whether this was the last step of the episode) \cite{Pardo2017-ht}. The state embedding was computed with a separate MPNN; the architecture of this model was identical to the encoder described in Section~\ref{encoder} except that it was not stochastic---only a single graph-level vector was computed. Thus, the value function is:
\begin{align}
V_T(s, y, t) &= g\left(\left[f_{\text{state}}(s), f_{\text{target}}(y), t_1, t_2\right]\right) \\
t_1 &= \frac{2(T-t)}{T}-1\\
t_2 &= \mathds{1}(t = T-1),
\end{align}
where $f_{\text{state}}$ is the state embedding MPNN, $f_{\text{target}}$ is the (stochastic) target embedding MPNN, $y$ is the input/target molecule, $T$ is the maximum number of steps in the episode (set to \num{20} in our experiments), and $t$ is the number of steps taken in the episode so far (\emph{i.e.}, the number of steps prior to but not including $s$). Note that (1) the target molecule embedding $f_{\text{target}}(y)$ was only sampled once per episode, not once per step; and (2) arbitrary points in the latent space can be decoded by replacing $f_{\text{target}}(y)$ with any embedding vector. The function $g$ was implemented as a fully connected neural network with a single hidden layer containing \num{256} ReLUs.

The reward is a numerical description of progress toward (or away from) a goal. In our case, the goal is reconstruction of an input molecule, and the reward measures the similarity between the input (target) molecule and the current state of the generator. In practice, we averaged over four different measures that capture different aspects of molecular similarity and provide complementary information to guide the generation process. The similarity measures that comprise the reward function can be divided into two groups:

\subsubsection{Fingerprint Similarity}

We used three types of molecular fingerprints, all computed with RDKit \cite{landrum2006rdkit}: extended-connectivity (ECFP), also known as ``circular'' or ``Morgan'' fingerprints, with $r=3$ \cite{Rogers2010-gp}; topological (path); and atom-pair \cite{Carhart1985-ip}. For each fingerprint, we computed the sparse (unhashed) Tversky similarity \cite{Horvath2013-mj} using the following $(\alpha,\beta)$ pairs: $\{(0.5,0.5),(0.95,0.05),(0.05,0.95)\}$. The resulting similarity values were then averaged to give a single value for each fingerprint type.

\subsubsection{Atom Count Similarity}

We computed the Tanimoto similarity across atom types:
\begin{equation}
T(A,B) = \frac{\sum_i \min \{\texttt{count}(A, i), \texttt{count}(B, i)\}}{\sum_i \max \{\texttt{count}(A, i),\texttt{count}(B, i)\}},
\end{equation}
where $A$ and $B$ are molecules, and $\texttt{count}(A, i)$ is the number of atoms in molecule $A$ with type $i$.

\subsection{Data}
\label{sec:data}

We used the QM9 dataset \cite{Ramakrishnan2014-lo}, which contains \num{133885} molecules with up to nine heavy atoms in the set \{C, N, O, F\}. The data was filtered to remove 1845 molecules containing atoms with formal charges (since our MDP cannot reproduce them) and randomly divided into ten folds. Models were trained on eight folds (\num{105773} examples), with one fold each held out for validation (``tuning''; \num{13154} examples) and testing (\num{13113} examples). Note that we did not repeat model training with different fold subsets; \emph{i.e.}, we did not perform cross-validation.

The QM9 dataset is suited for experimentation since its molecules are relatively small. However, they tend to be relatively complex since the dataset attempts to exhaustively enumerate all possible structures of that size and containing that set of atoms. Additionally, fingerprint-based similarity metrics can become brittle when applied to small molecules, since they do not set as many bits in the fingerprint---see \figurename~A1 in the Supplemental Material for examples of QM9-sized molecule pairs with different similarity values.

\subsection{Model Training}

Models were implemented with TensorFlow \cite{tensorflow2015-whitepaper} and trained with the Adam optimizer \cite{Kingma2014-js} with $\beta_1=0.9$ and $\beta_2=0.999$. Training proceeded for $\sim$\num{10}~M steps with a batch size of \num{128}; the learning rate was smoothly decayed from $10^{-5}$ with a decay rate of \num{0.99} every \num{100000} steps.

\subsubsection{Training Loss}

The training loss had the form:
\begin{equation}
\mathcal{L}(s, y, \mu, \Sigma)=H\left(\mathcal{L}_{\text{TD}}(s, y)\right) + \lambda \mathcal{L}_{\text{KL}}\left(\mu, \Sigma\right).
\end{equation}
$\mathcal{L}_{\text{TD}}(s)$ is the temporal difference loss used to train the decoder:
\begin{align}
\mathcal{L}_{\text{TD}}(s, y) &= V_T(s, y, t) - \hat{V}_T(s, y, t) \\
\hat{V}_T(s, y, t) &= R(s,y) + \gamma \max_{s'} V_T(s', y, t + 1),
\label{eq:value}
\end{align}
where $R$ is the reward function (see Section~\ref{sec:decoder}), $\gamma$ is the discount factor (we used \num{0.99}), and $H$ is the Huber loss function with $\delta=1$ \cite{Huber1964-ri}. Additionally, the weight on this loss for each example was adjusted such that terminal and non-terminal states had approximately the same ``mass'' in each batch.

$\mathcal{L}_{\text{KL}}(h)$ is the Kullback--Leibler divergence between a multivariate unit Gaussian prior and the variational distribution---parameterized by the predicted mean vector $\mu$ and diagonal covariance matrix $\Sigma$---from which the target molecule embedding is sampled. The hyperparameter $\lambda$ controls the relative contribution of this term to the total loss ($\lambda=10^{-5}$ in all of the experiments reported here).

\subsubsection{Experience Replay}

The states $s$ used in training were uniformly sampled from an experience replay buffer \cite{Mnih2015-ye}. The buffer was updated in each training step by sampling a batch of molecules from the training data and generating two episodes for each molecule:
\begin{enumerate}
\item The current value function approximation was used to run an episode with $\epsilon$-greedy action selection. $\epsilon$ started at \num{1.0} and was smoothly decayed during training with a decay rate of \num{0.95} every \num{10000} steps.
\item An idealized episode was created to reconstruct the target molecule (see Section~A of the Supplemental Material for details). This ensured that the replay buffer included examples of productive steps.
\end{enumerate}
The replay buffer stored states, rewards, actions, and the number of steps taken; each \num{20}-step episode created \num{20} examples for the buffer. At the beginning of model training, the buffer was populated to contain at least \num{1000} examples by running several batches without any training. When the buffer reached the maximum size of \num{10000} examples, the oldest examples were discarded to make room for new examples.

Each training step used a batch size of \num{8} to generate episode data for the replay buffer (corresponding to \num{320} buffer entries) and a batch size of \num{128} for sampling from the buffer to compute the training loss.

\section{Results and Discussion}

\subsection{Reconstruction}

After training an RL-VAE model on a subset of the QM9 data set, we evaluated reconstruction performance on the held-out tune and test sets, each containing $\sim$\num{13000} molecules (see Section~\ref{sec:data}). Each molecule was encoded as a distribution in the latent space, and a single embedding was sampled from this distribution and decoded.

We computed three metrics: (1) reconstruction accuracy, measured by canonical SMILES equivalence (ignoring stereochemistry, which is not predicted by the decoder); (2) Tanimoto similarity between input and output molecules; and (3) MDP edit distance, measured as the number of MDP steps required to reach the decoded molecule from the target molecule (calculated by brute-force search; see Section~B of the Supplemental Material for details). We also verified that a greedy model with discount factor $\gamma=0$ performed significantly worse.
Reconstruction accuracy is shown in \tablename~\ref{table:reconstruction}.
Tanimoto similarity between input and output molecules fell off sharply if molecules were not perfectly reconstructed (\figurename~A2 and \figurename~A3 in the Supplemental Material), suggesting that MDP edit distance is a more useful metric for molecules of this size (see Section~\ref{sec:data}). \figurename~A4 in the Supplemental Material shows the distribution of MDP edit distances for the non-greedy model.

\begin{table}[tb]
\caption{Reconstruction accuracy for various models. The random walk model chooses actions randomly.
The symbol $\gamma$ is the discount factor in Eq.~\ref{eq:value}; $\gamma=0$ only considers immediate rewards. Details on JT-VAE and GVAE training and evaluation are given in Section~D of the Supplemental Material.}
\label{table:reconstruction}
\vskip 0.15in
\begin{center}
\begin{small}
\begin{sc}
\begin{tabular}{lSS}
\toprule
 & \text{Tune} & \text{Test} \\
\midrule
Random Walk & 0.00 & 0.00 \\
RL-VAE ($\gamma=0$) & 0.03  & 0.03 \\
RL-VAE ($\gamma=0.99$) & 0.66 & 0.67 \\
JT-VAE~\textup{\cite{Jin2018-rz}} & 0.74 & 0.74 \\
GVAE~\textup{\cite{Kusner2017-dr}} & 0.51 & 0.51 \\
\bottomrule
\end{tabular}
\end{sc}
\end{small}
\end{center}
\vskip -0.1in
\end{table}

\subsection{Sampling}

We randomly chose two orthogonal directions that we used to explore the latent space, decoding molecules along the way. As we suspect is the case for most explorations of this type, the results are highly sensitive to the choice of directions, the starting molecule, the distance travelled in each direction, and the step size. \figurename~A5 in the Supplemental Material shows an example of such an exploration that reveals some local structure, although the decoded molecules do not exhibit very smooth transitions (\emph{i.e.}, transitions often change more than one aspect of a structure).

\begin{figure}[tb]
\begin{center}
\centerline{\includegraphics[width=\linewidth]{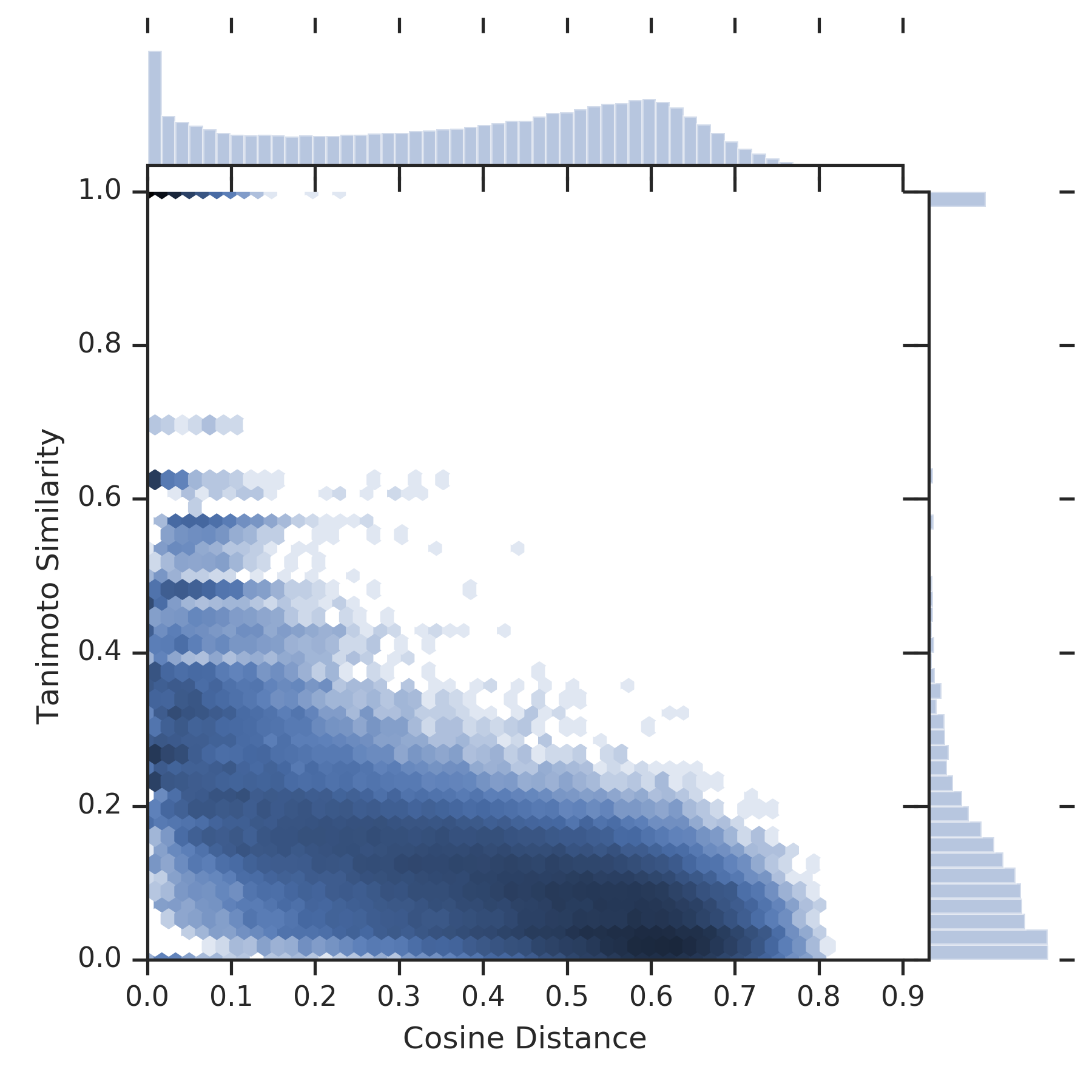}}
\caption{Tanimoto similarity between molecules decoded from original and perturbed embeddings as a function of cosine distance in the latent space (combined data from ten different original embeddings). The density color map is logarithmic.}
\label{fig:distance}
\end{center}
\vskip -0.25in
\end{figure}

As an attempt to more rigorously investigate the smoothness of the embedding space, we measured the similarity of decoded compounds as a function of cosine distance in the latent space (see Section~C of the Supplemental Material for details). As shown in \figurename~\ref{fig:distance}, we observed that the similarity of the decoded molecules tends to decrease as the cosine distance increases, suggesting that the embedding space around these starting points has relatively smooth local structure (see the Supplemental Material for additional plots for each starting point).

\section{Conclusion and Future Work}

This paper describes a graph-to-graph autoencoder with an unusual training signal: temporal difference (TD) errors for $Q$-function predictions. Our preliminary results suggest that RL-VAE is a simple and effective approach that bridges the gap between embedding-based methods and generative models for molecular generation and optimization.


As this is a workshop paper describing preliminary work, we anticipate several avenues for future work, including: (1) additional hyperparameter optimization; (2) training on larger and more relevant data sets, such as ChEMBL \cite{Gaulton2012-nw}; (3) evaluation on optimization tasks that are relevant for drug discovery---in particular, this excludes properties like logP or QED that are only useful as baselines (see \citet{Zhou2018-yc}); and (4) Bayesian optimization as demonstrated by \citet{Gomez-Bombarelli2018-yy}. 




\bibliography{references}
\bibliographystyle{icml2019}

\end{document}


\twocolumn[
\icmltitle{Decoding Molecular Graph Embeddings with Reinforcement Learning\\ \textsc{Supplemental Material}}



\icmlsetsymbol{equal}{*}

\begin{icmlauthorlist}
\icmlauthor{Steven Kearnes}{google}
\icmlauthor{Li Li}{google}
\icmlauthor{Patrick Riley}{google}
\end{icmlauthorlist}

\icmlaffiliation{google}{Google Research, Mountain View, California, USA}

\icmlcorrespondingauthor{Steven Kearnes}{kearnes@google.com}

\icmlkeywords{Machine Learning, ICML}

\vskip 0.3in
]



\printAffiliationsAndNotice{}  
\renewcommand{\thefigure}{A\arabic{figure}}
\renewcommand{\thetable}{A\arabic{table}}
\renewcommand\thesection{\Alph{section}}

\appendix

\begin{figure*}[ht]
\centering
\begin{subfigure}[b]{0.3\textwidth}
\includegraphics{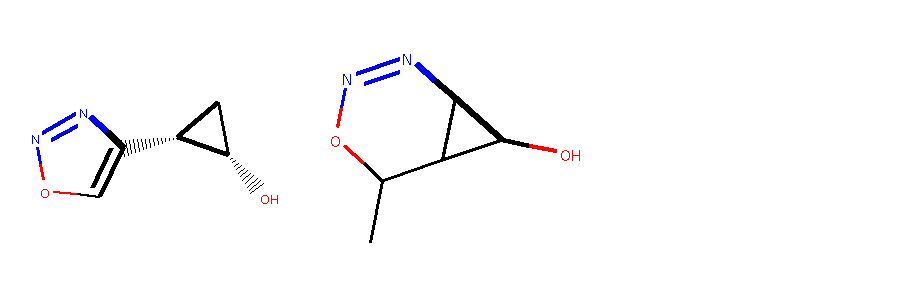}
\caption{\num{0.13}}
\end{subfigure}
\begin{subfigure}[b]{0.3\textwidth}
\includegraphics{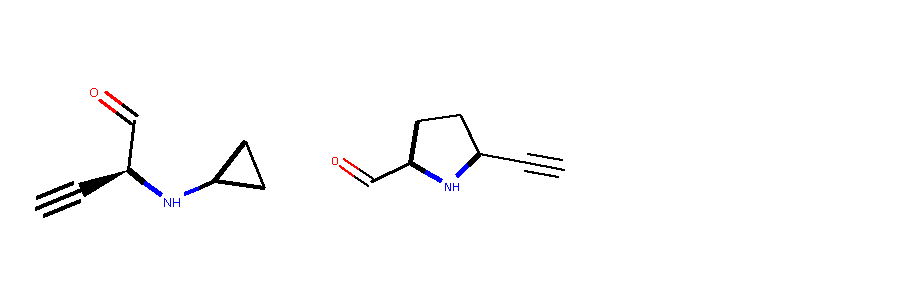}
\caption{\num{0.25}}
\end{subfigure}
\begin{subfigure}[b]{0.3\textwidth}
\includegraphics{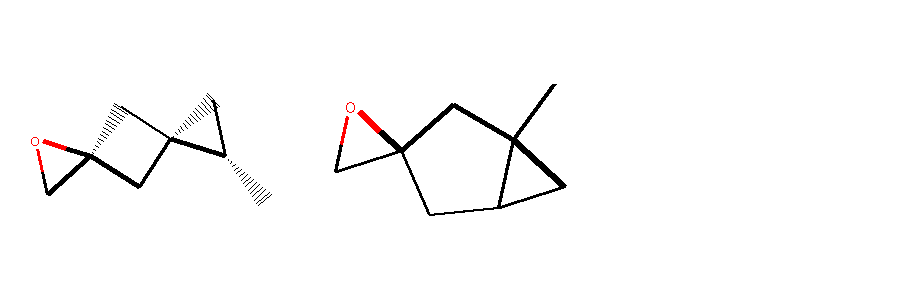}
\caption{\num{0.38}}
\end{subfigure}
\begin{subfigure}[b]{0.3\textwidth}
\includegraphics{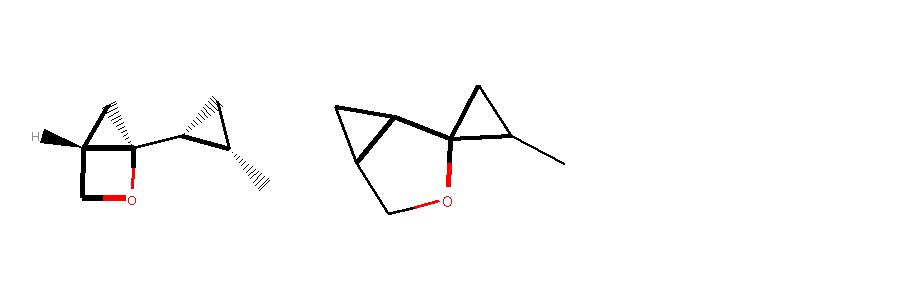}
\caption{\num{0.47}}
\end{subfigure}
\begin{subfigure}[b]{0.3\textwidth}
\includegraphics{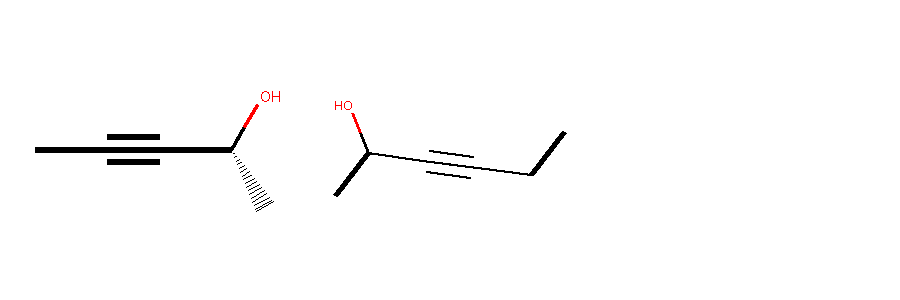}
\caption{\num{0.48}}
\end{subfigure}
\begin{subfigure}[b]{0.3\textwidth}
\includegraphics{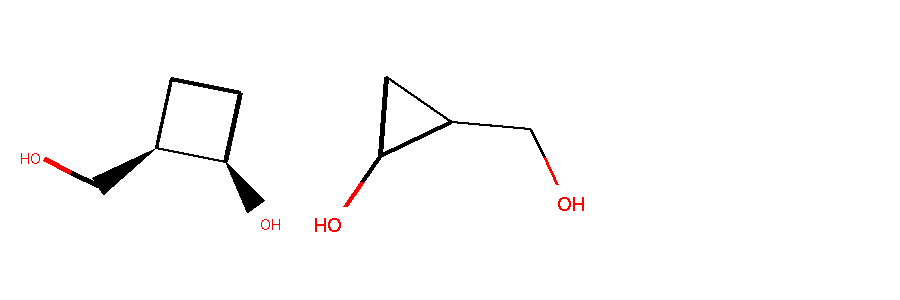}
\caption{\num{0.55}}
\end{subfigure}
\begin{subfigure}[b]{0.3\textwidth}
\includegraphics{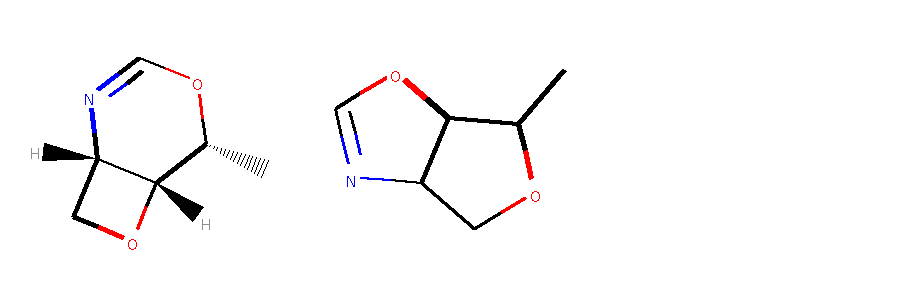}
\caption{\num{0.57}}
\end{subfigure}
\begin{subfigure}[b]{0.3\textwidth}
\includegraphics{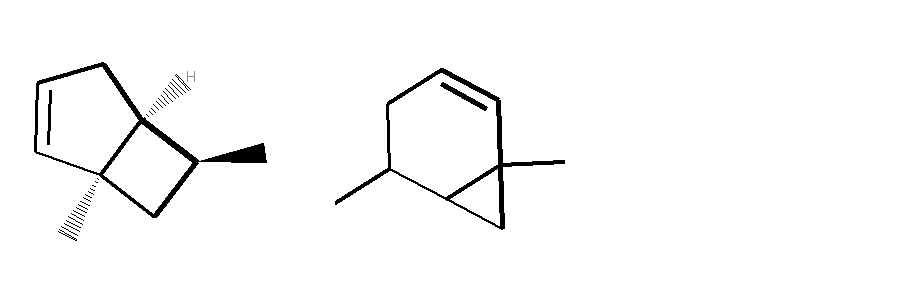}
\caption{\num{0.59}}
\end{subfigure}
\begin{subfigure}[b]{0.3\textwidth}
\includegraphics{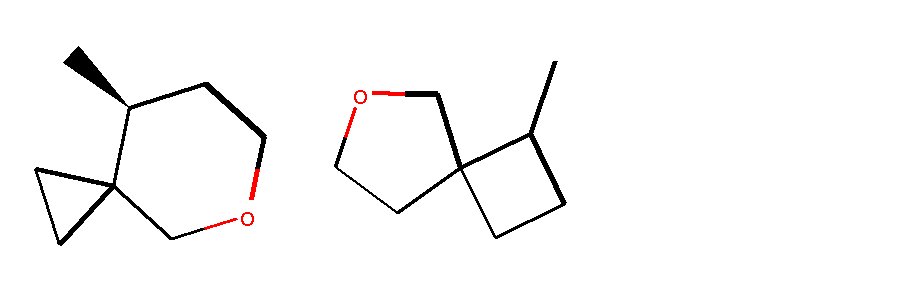}
\caption{\num{0.63}}
\end{subfigure}
\begin{subfigure}[b]{0.3\textwidth}
\includegraphics{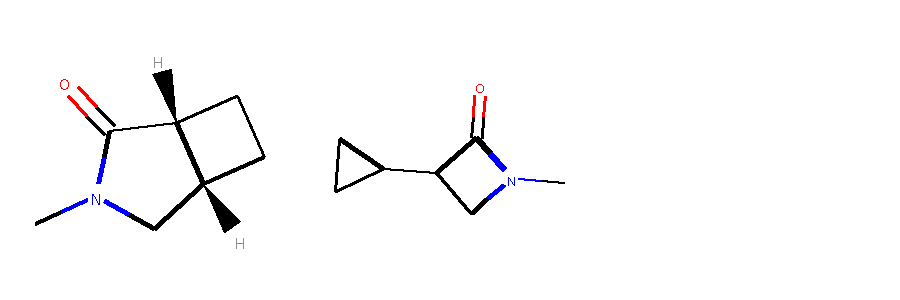}
\caption{\num{0.64}}
\end{subfigure}
\begin{subfigure}[b]{0.3\textwidth}
\includegraphics{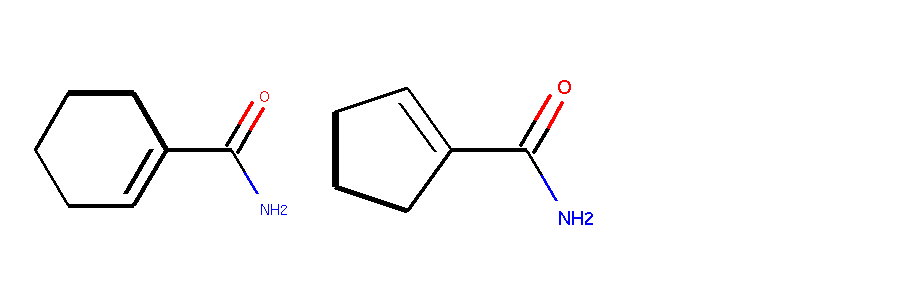}
\caption{\num{0.68}}
\end{subfigure}
\begin{subfigure}[b]{0.3\textwidth}
\includegraphics{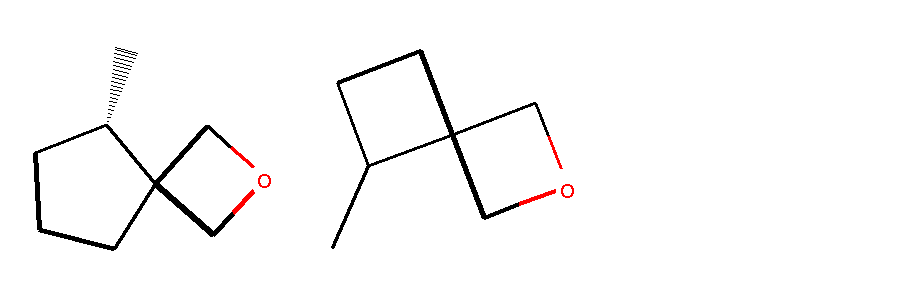}
\caption{\num{0.69}}
\end{subfigure}
\begin{subfigure}[b]{0.3\textwidth}
\includegraphics{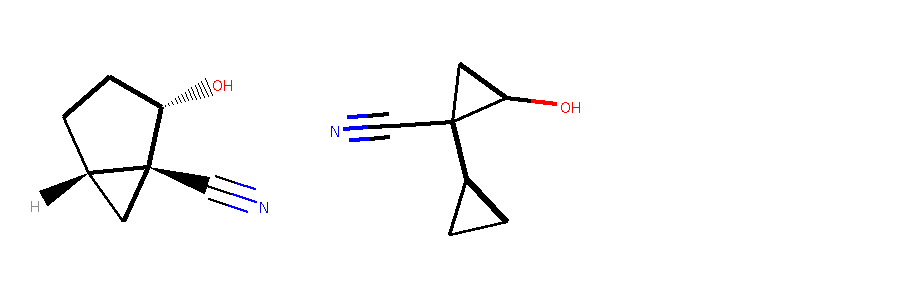}
\caption{\num{0.70}}
\end{subfigure}
\begin{subfigure}[b]{0.3\textwidth}
\includegraphics{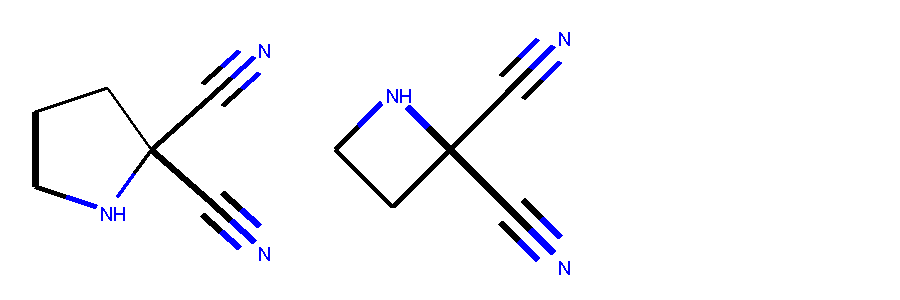}
\caption{\num{0.71}}
\end{subfigure}
\begin{subfigure}[b]{0.3\textwidth}
\includegraphics{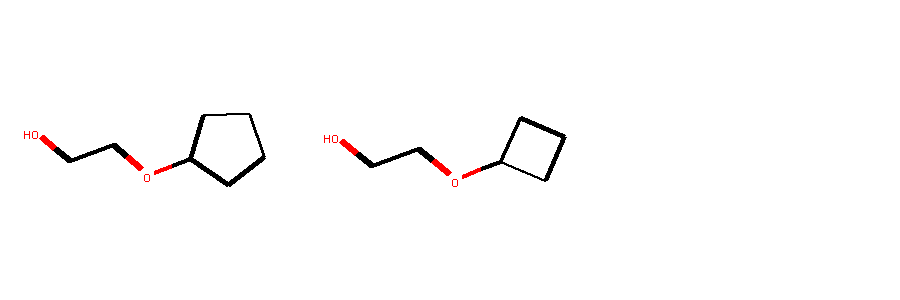}
\caption{\num{0.75}}
\end{subfigure}
\begin{subfigure}[b]{0.3\textwidth}
\includegraphics{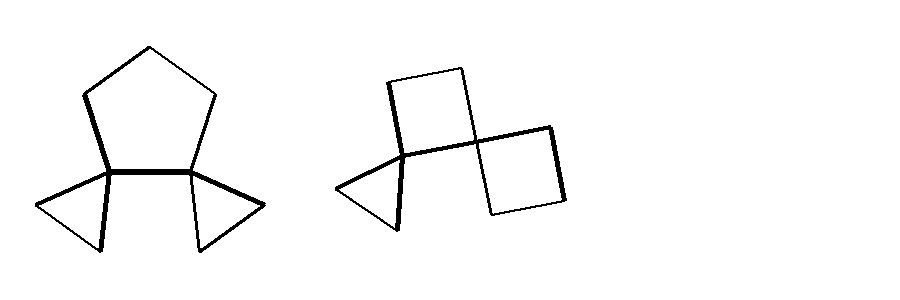}
\caption{\num{0.77}}
\end{subfigure}
\begin{subfigure}[b]{0.3\textwidth}
\includegraphics{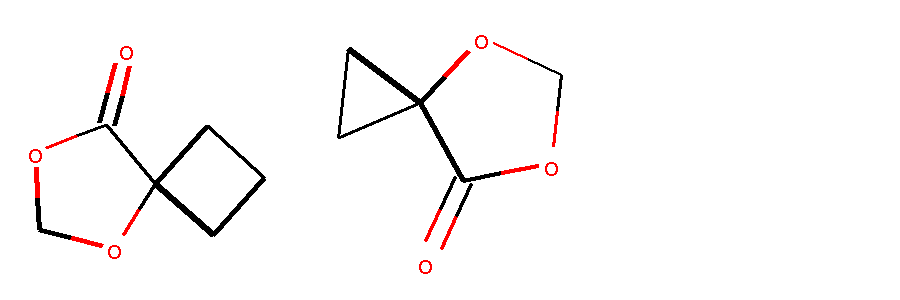}
\caption{\num{0.82}}
\end{subfigure}
\begin{subfigure}[b]{0.3\textwidth}
\includegraphics{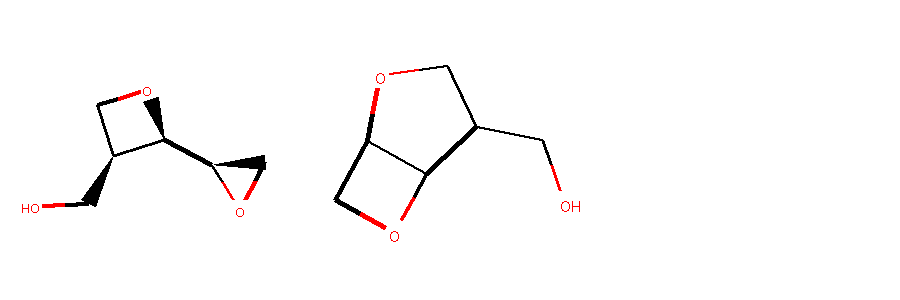}
\caption{\num{0.90}}
\end{subfigure}
\caption{Tanimoto similarity for various molecule pairs. In each pair, the molecule on the left is taken from the QM9 tune set. Similarity is computed on sparse Morgan fingerprints with radius \num{3}; note that stereochemistry is ignored. The interpretation of similarity values varies with the size of the molecule as the total number of bits set in the fingerprints changes.}
\label{fig:sim}
\end{figure*}

\section{Idealized episodes}

To construct idealized episodes, molecules were broken into their constituent atoms and assembled one step at a time. When a new atom was added to the graph, bonds to existing atoms were added as additional steps (\emph{i.e.}, atom and bond additions were interleaved in the episode). The first atom added to the graph corresponded to the first atom in the list of atoms returned by \texttt{RDKit}; subsequent atoms were added if they had a bond to at least one existing atom in the graph.

\section{Brute-force search}

MDP edit distance was calculated by brute-force search from molecule $A$ to molecule $B$. Actions allowed from molecule $A$ were exhaustively expanded until molecule $B$ or a maximum step limit was reached. The MDP used for the brute-force search allowed a more flexible set of actions; in particular: (1) bond removal/promotion was allowed, (2) states were Kekulized so that bonds in aromatic systems could be modified, (3) rings of any size were allowed, (4) the ``no modification'' action was not allowed, and (5) bonds between ring atoms were allowed. This set of actions resulted a very flexible MDP that helped match the calculated edit distance to intuitive notions of distance.

\section{Latent space embedding perturbations}

\begin{figure*}[tb]
\begin{center}
\begin{subfigure}[b]{0.24\textwidth}
\includegraphics[width=\linewidth]{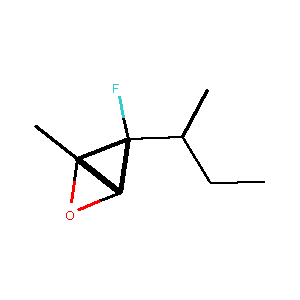}
\caption{\texttt{CCC(C)C1(F)C2OC21C}}
\end{subfigure}
\begin{subfigure}[b]{0.24\textwidth}
\includegraphics[width=\linewidth]{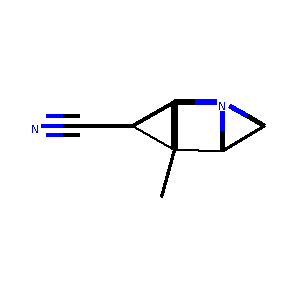}
\caption{\texttt{CC12C(C\#N)C1N1CC12}}
\end{subfigure}
\begin{subfigure}[b]{0.24\textwidth}
\includegraphics[width=\linewidth]{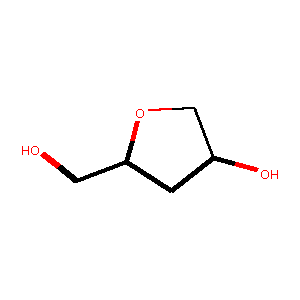}
\caption{\texttt{OCC1CC(O)CO1}}
\end{subfigure}
\begin{subfigure}[b]{0.24\textwidth}
\includegraphics[width=\linewidth]{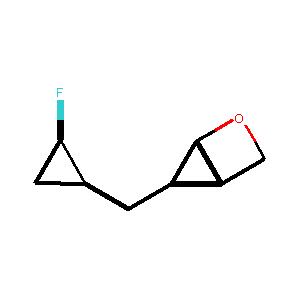}
\caption{\texttt{FC1CC1CC1C2COC21}}
\end{subfigure}
\begin{subfigure}[b]{0.24\textwidth}
\includegraphics[width=\linewidth]{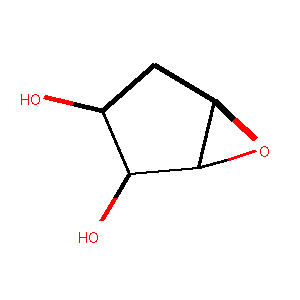}
\caption{\texttt{OC1CC2OC2C1O}}
\end{subfigure}
\begin{subfigure}[b]{0.24\textwidth}
\includegraphics[width=\linewidth]{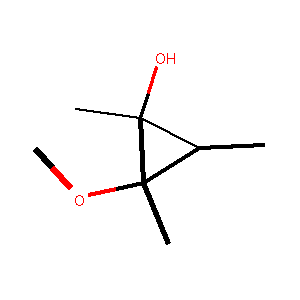}
\caption{\texttt{COC1(C)C(C)C1(C)O}}
\end{subfigure}
\begin{subfigure}[b]{0.24\textwidth}
\includegraphics[width=\linewidth]{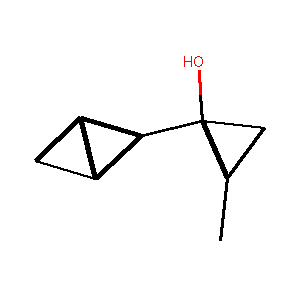}
\caption{\texttt{CC1CC1(O)C1C2CC21}}
\end{subfigure}
\begin{subfigure}[b]{0.24\textwidth}
\includegraphics[width=\linewidth]{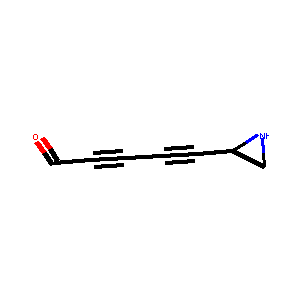}
\caption{\texttt{O=CC\#CC\#CC1CN1}}
\end{subfigure}
\begin{subfigure}[b]{0.24\textwidth}
\includegraphics[width=\linewidth]{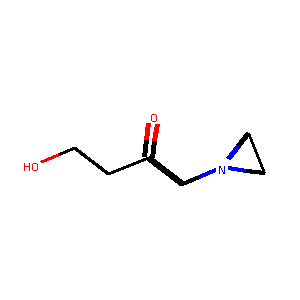}
\caption{\texttt{O=C(CCO)CN1CC1}}
\end{subfigure}
\begin{subfigure}[b]{0.24\textwidth}
\includegraphics[width=\linewidth]{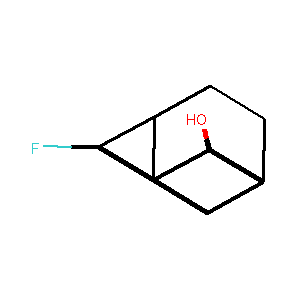}
\caption{\texttt{OC1C2CCC3C(F)C13C2}}
\end{subfigure}
\caption{Molecules decoded from original embeddings used for latent space exploration. A single embedding that decoded to each molecule was used as the starting point for the explorations described in Section~3.2.}
\label{fig:sampled_mols}
\end{center}
\vskip -0.2in
\end{figure*}

For an array of scaling factors spanning $[-5,5]$ in $0.1$ intervals (excluding $0.0$), we performed the following procedure: Starting with a single embedding, we uniformly sampled a \num{256}-dimensional perturbation vector in the range $[0, 1)$, and decoded the perturbed embedding. This was repeated \num{100} times for each scaling factor, and the entire process was again repeated for ten different starting embeddings (representing ten different starting molecules; \figurename~\ref{fig:sampled_mols}), for a total of \num{100000} perturbations. Cosine and Euclidean distances were calculated between the original embedding and the perturbed embedding; Tanimoto similarities between the input molecule and the molecule decoded from the perturbed embedding were calculated using sparse Morgan fingerprints with radius~\num{3}.

\begin{figure*}[ht]
\begin{center}
\begin{subfigure}[b]{0.49\textwidth}
\includegraphics[width=\linewidth]{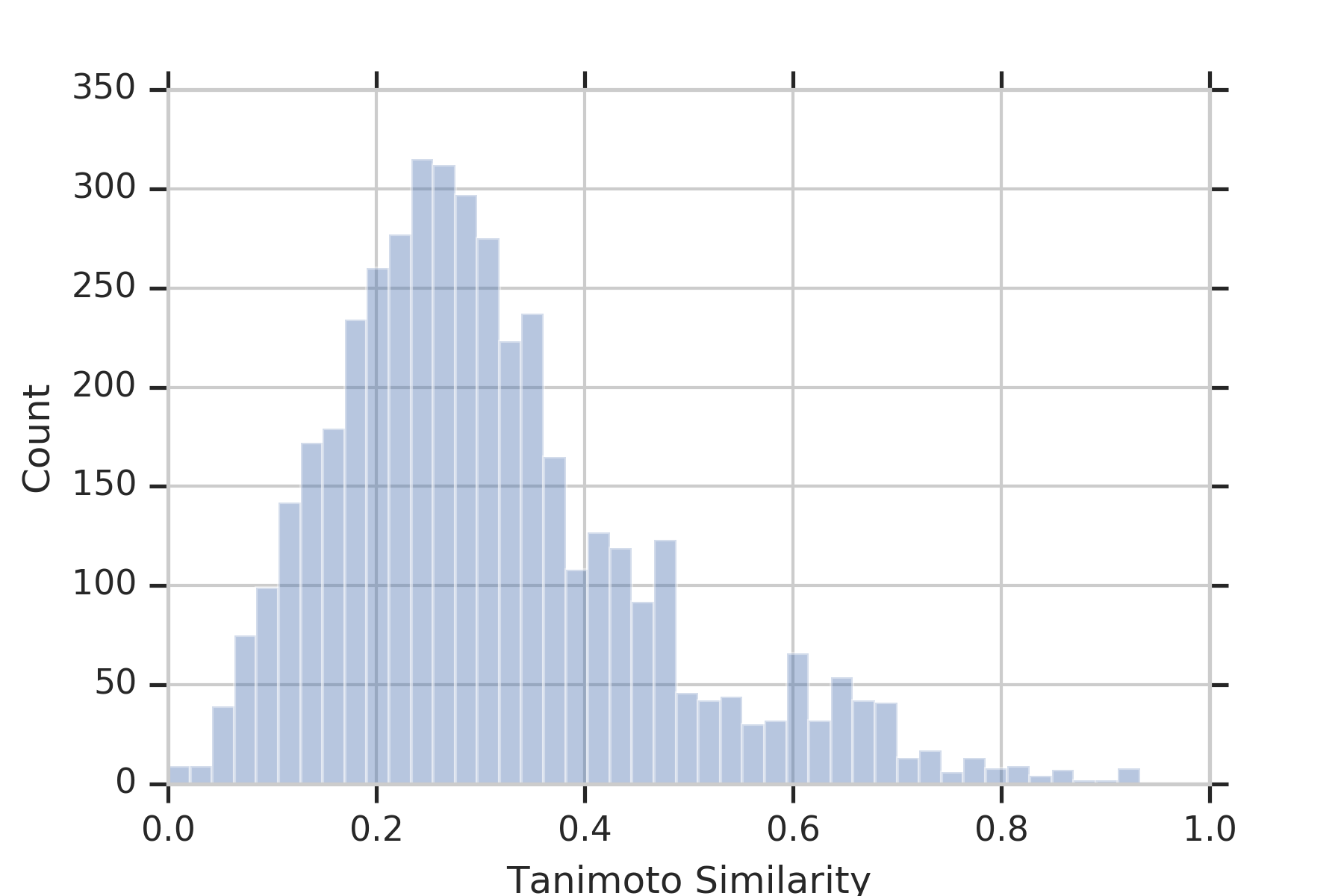}
\caption{Tune}
\end{subfigure}
\begin{subfigure}[b]{0.49\textwidth}
\includegraphics[width=\linewidth]{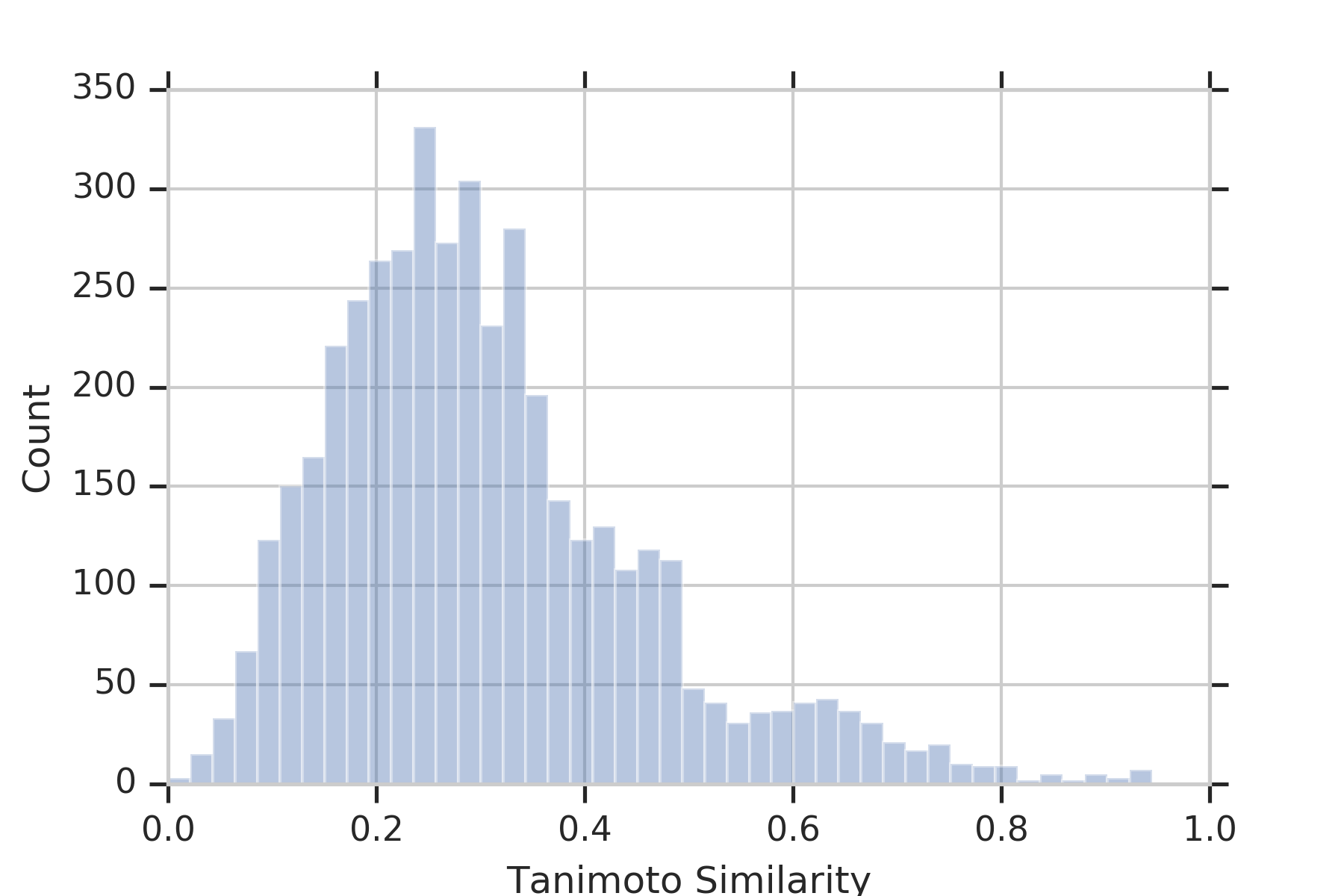}
\caption{Test}
\end{subfigure}
\caption{Tanimoto similarity of input and output molecules for $\gamma=0.99$. Identical molecules ($\text{similarity}=1$) are excluded for clarity. Similarity was computed on sparse Morgan fingerprints with radius \num{3}.}
\label{fig:explore}
\end{center}
\vskip -0.2in
\end{figure*}

\begin{figure*}[ht]
\begin{center}
\begin{subfigure}[b]{0.49\textwidth}
\includegraphics[width=\linewidth]{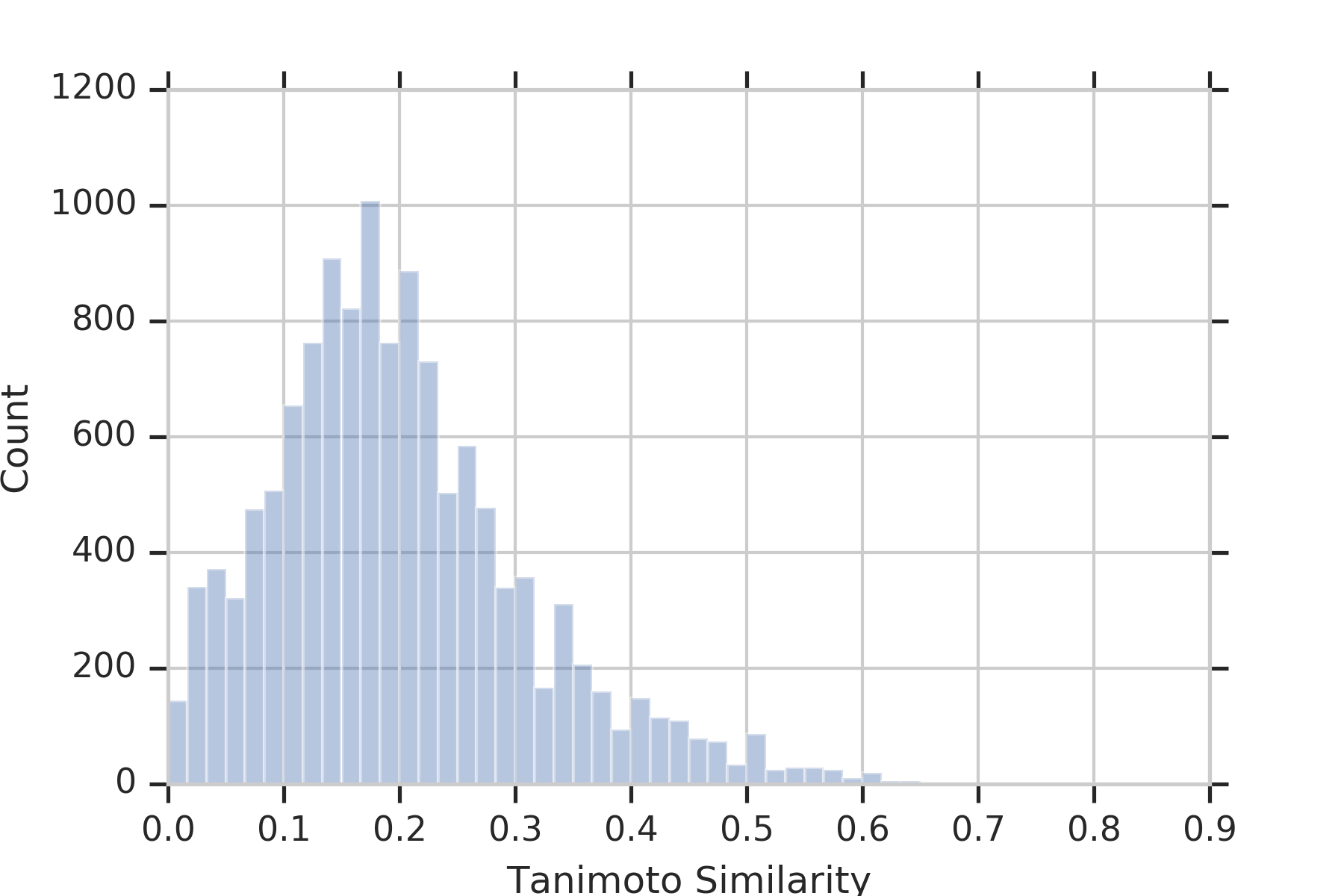}
\caption{Tune}
\end{subfigure}
\begin{subfigure}[b]{0.49\textwidth}
\includegraphics[width=\linewidth]{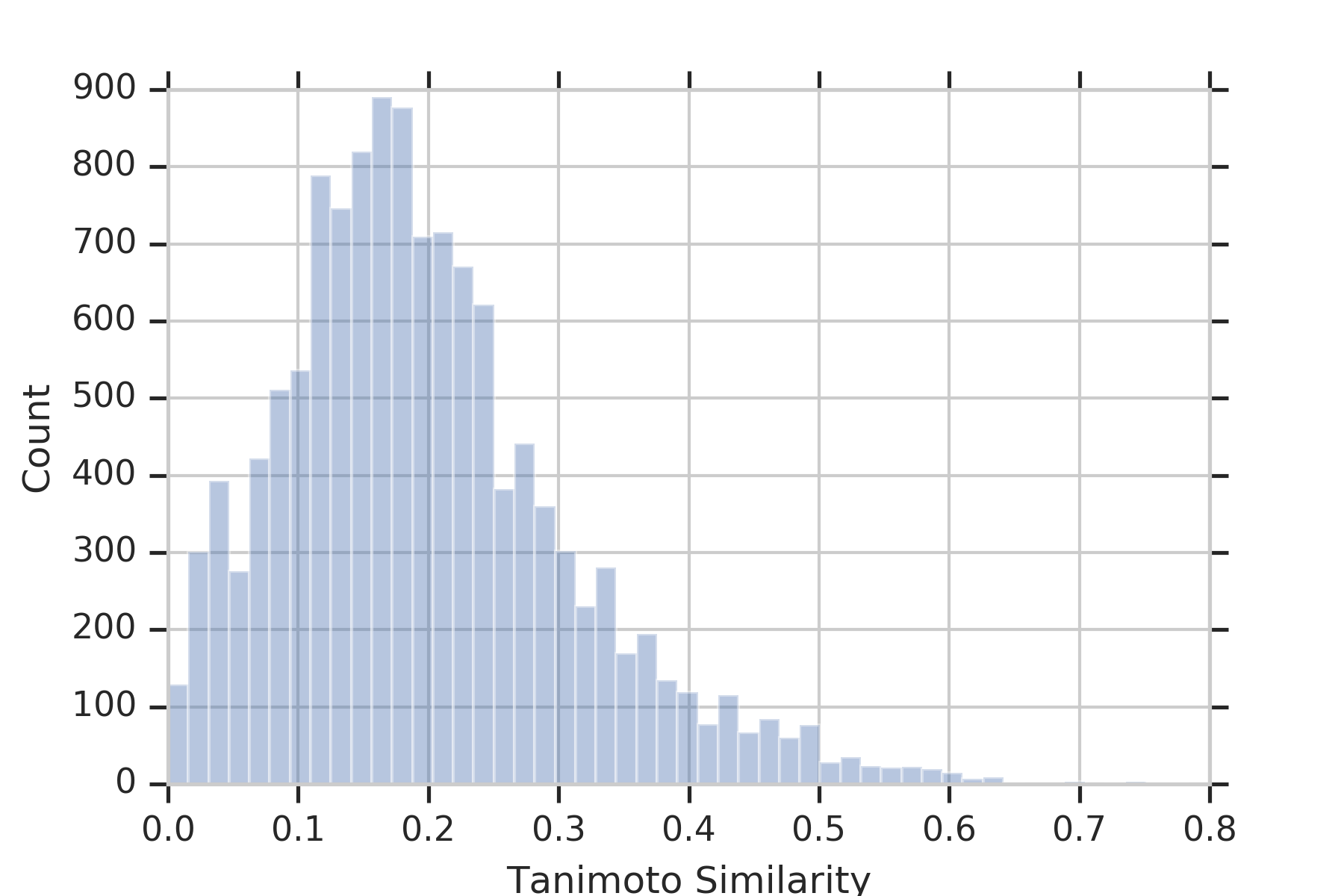}
\caption{Test}
\end{subfigure}
\caption{Tanimoto similarity of input and output molecules for $\gamma=0$. Identical molecules ($\text{similarity}=1$) are excluded for clarity. Similarity was computed on sparse Morgan fingerprints with radius \num{3}.}
\label{fig:explore}
\end{center}
\vskip -0.2in
\end{figure*}

\begin{figure*}[ht]
\begin{center}
\begin{subfigure}[b]{0.49\textwidth}
\includegraphics[width=\linewidth]{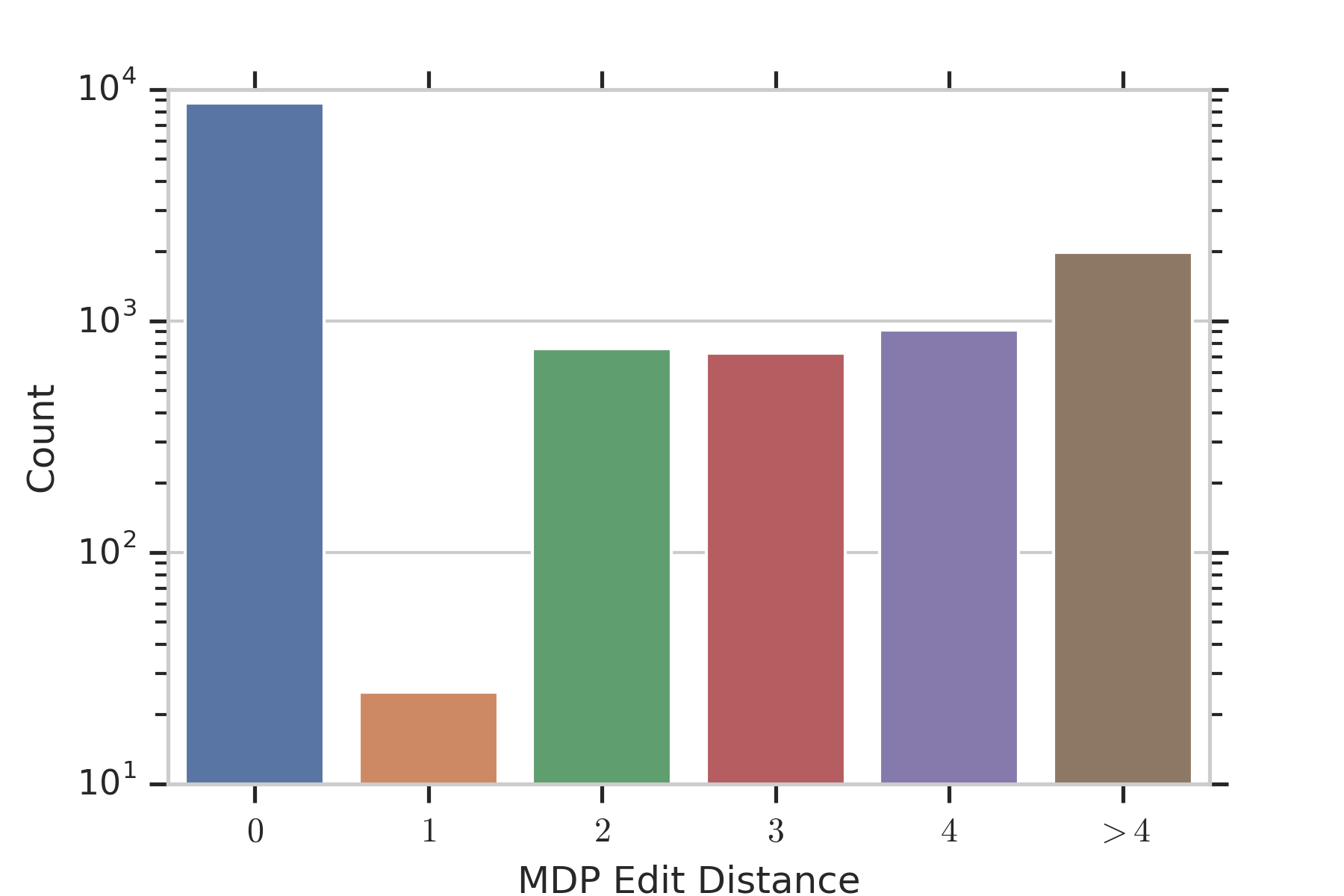}
\caption{Tune}
\end{subfigure}
\begin{subfigure}[b]{0.49\textwidth}
\includegraphics[width=\linewidth]{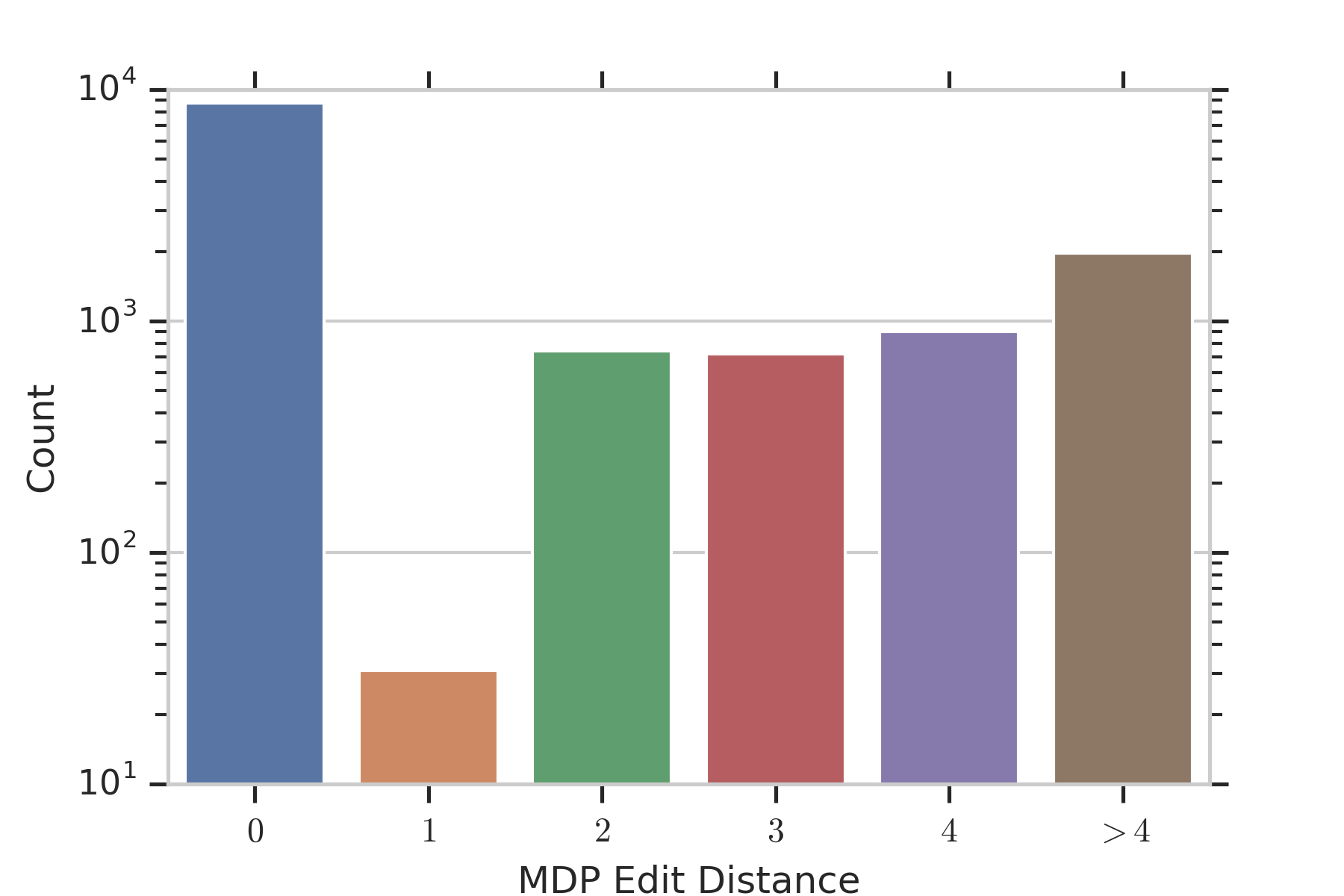}
\caption{Test}
\end{subfigure}
\caption{MDP edit distance between input and output molecules for $\gamma=0.99$.}
\label{fig:explore}
\end{center}
\vskip -0.2in
\end{figure*}

\begin{figure*}[ht]
\begin{center}
\centerline{\includegraphics[width=\linewidth]{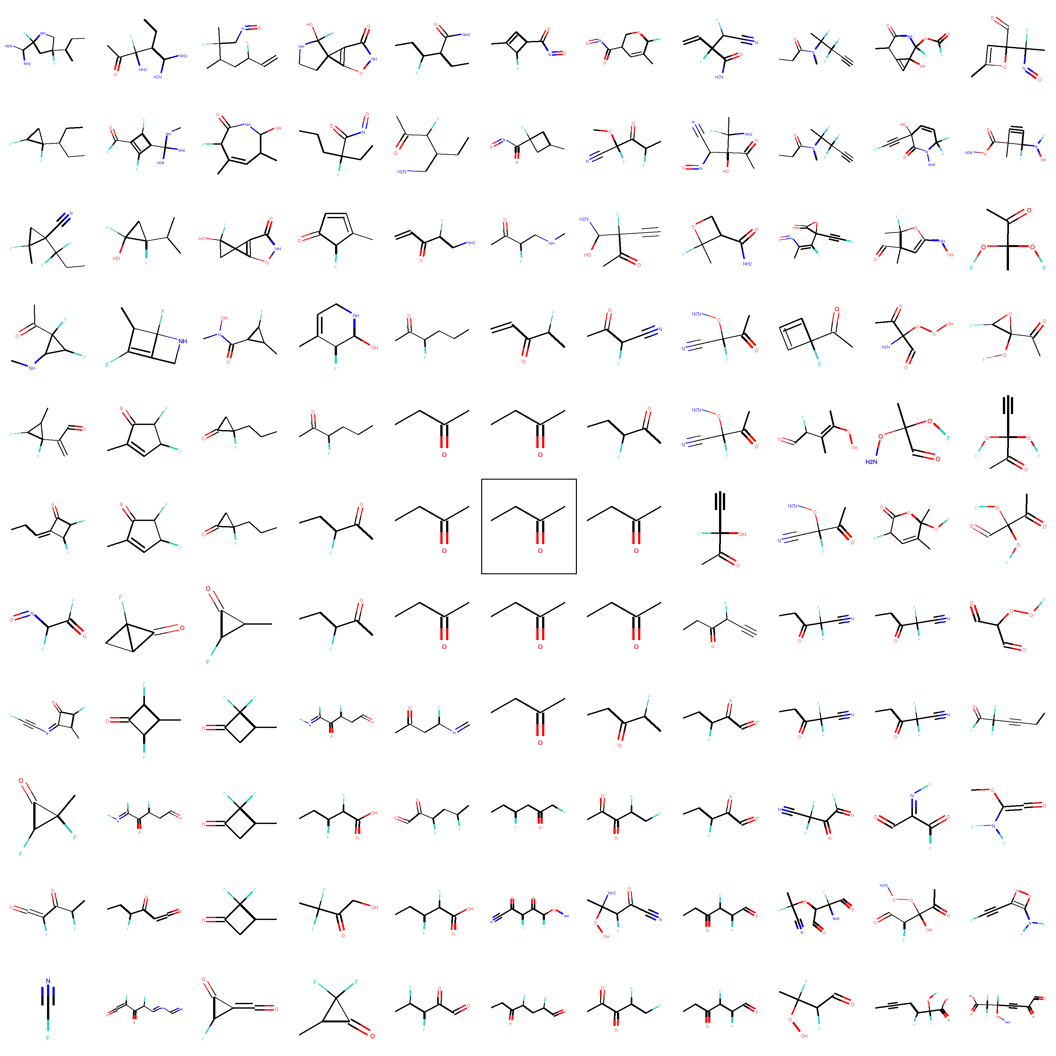}}
\caption{Exploration of the embedding space in two random orthogonal directions. The starting molecule (boxed) was \texttt{CCC(C)=O}. The edges of the image correspond to embeddings at $\pm20$x a unit vector in that direction; intermediate points were sampled at intervals of \num{4}x.}
\label{fig:explore}
\end{center}
\vskip -0.2in
\end{figure*}

\begin{figure*}[tb]
\begin{center}
\centerline{\includegraphics[width=\linewidth]{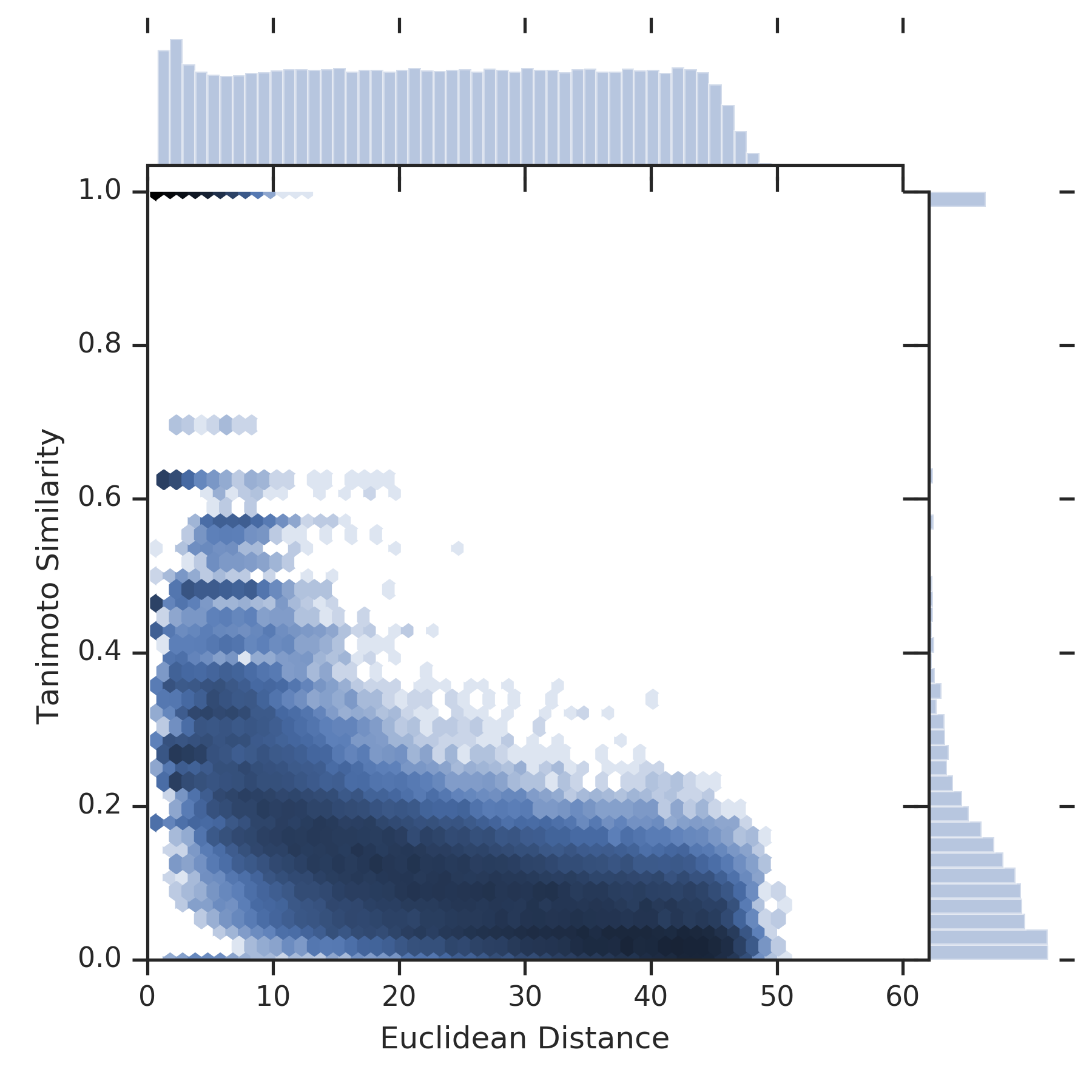}}
\caption{Tanimoto similarity between molecules decoded from original and perturbed embeddings as a function of cosine distance in the latent space (combined data from ten different original embeddings) as a function of Euclidean distance in the latent space, starting from ten different embedding locations. The density color map is logarithmic. Similarity was computed on sparse Morgan fingerprints with radius \num{3}.}
\label{fig:distance}
\end{center}
\vskip -0.2in
\end{figure*}

\begin{figure*}[tb]
\begin{center}
\begin{subfigure}[b]{0.24\textwidth}
\includegraphics[width=\linewidth]{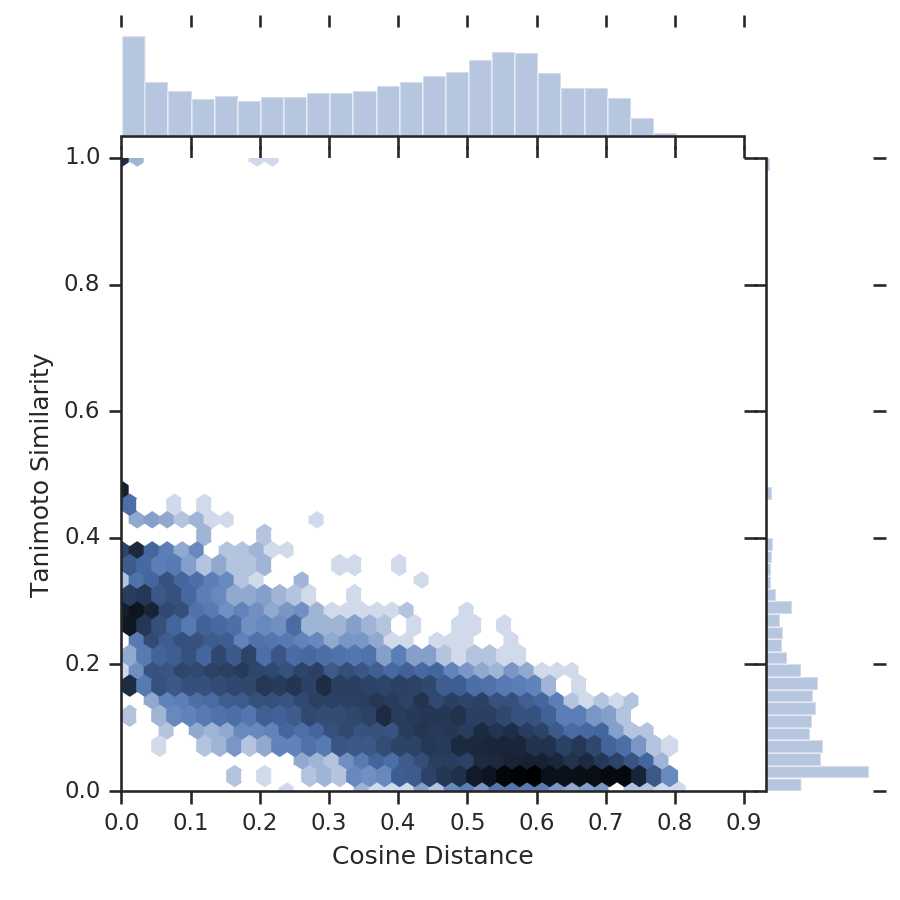}
\caption{\texttt{CCC(C)C1(F)C2OC21C}}
\end{subfigure}
\begin{subfigure}[b]{0.24\textwidth}
\includegraphics[width=\linewidth]{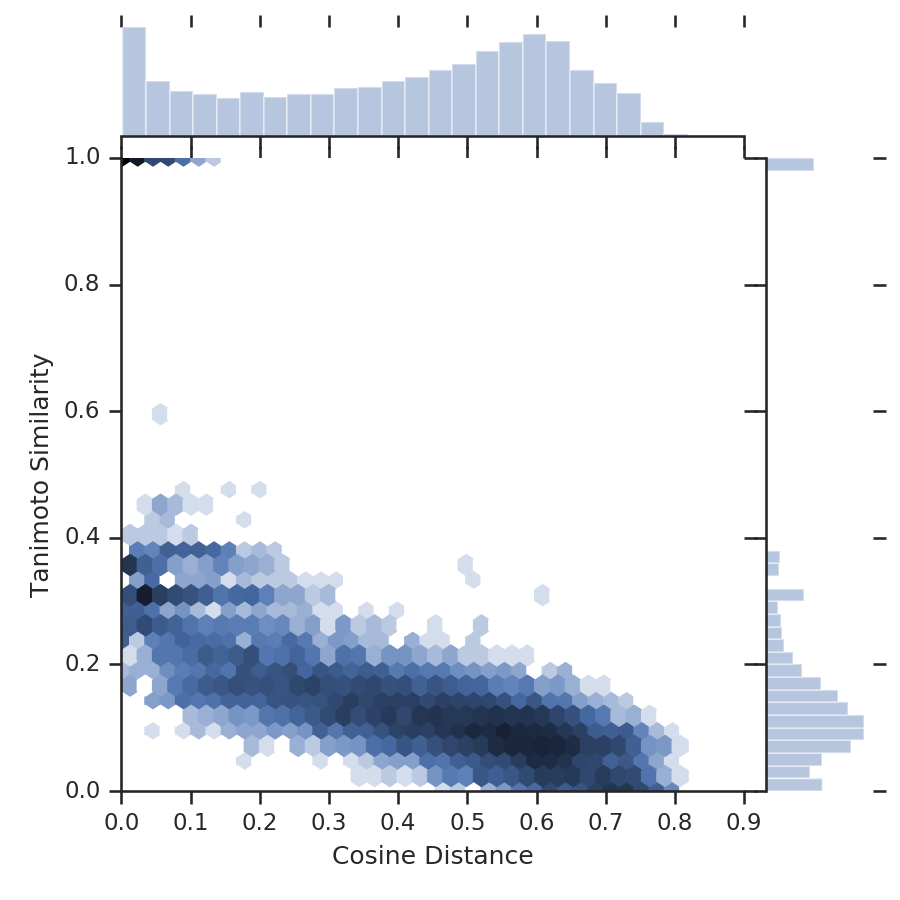}
\caption{\texttt{CC12C(C\#N)C1N1CC12}}
\end{subfigure}
\begin{subfigure}[b]{0.24\textwidth}
\includegraphics[width=\linewidth]{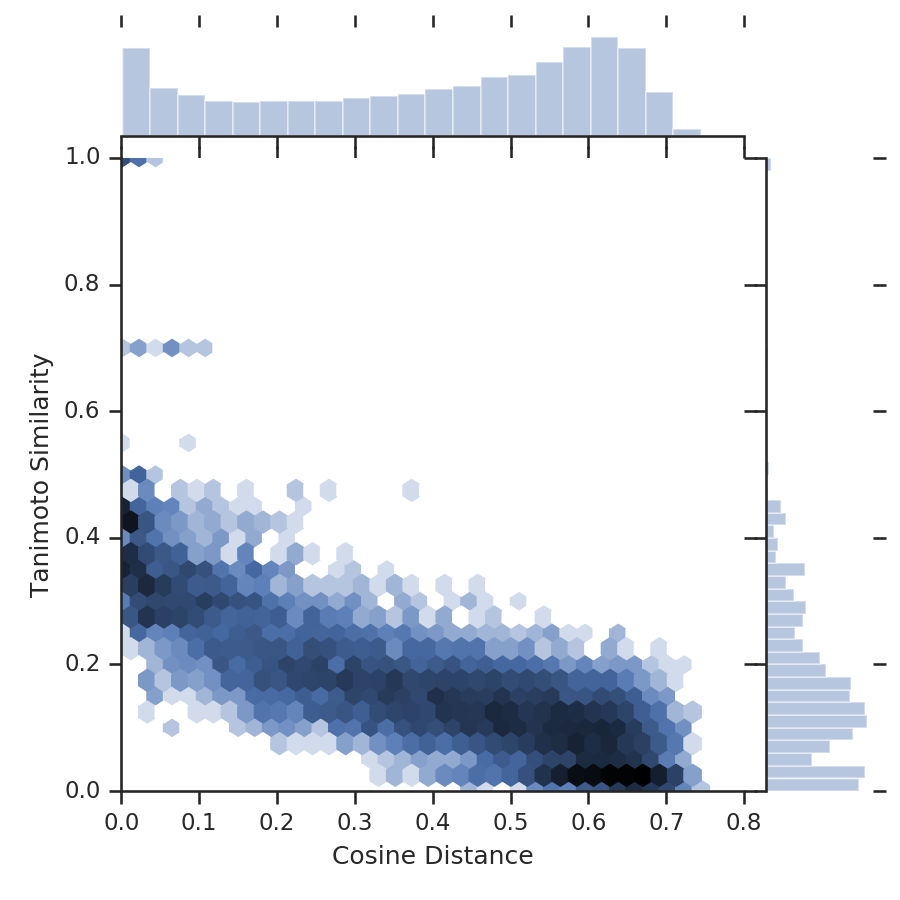}
\caption{\texttt{OCC1CC(O)CO1}}
\end{subfigure}
\begin{subfigure}[b]{0.24\textwidth}
\includegraphics[width=\linewidth]{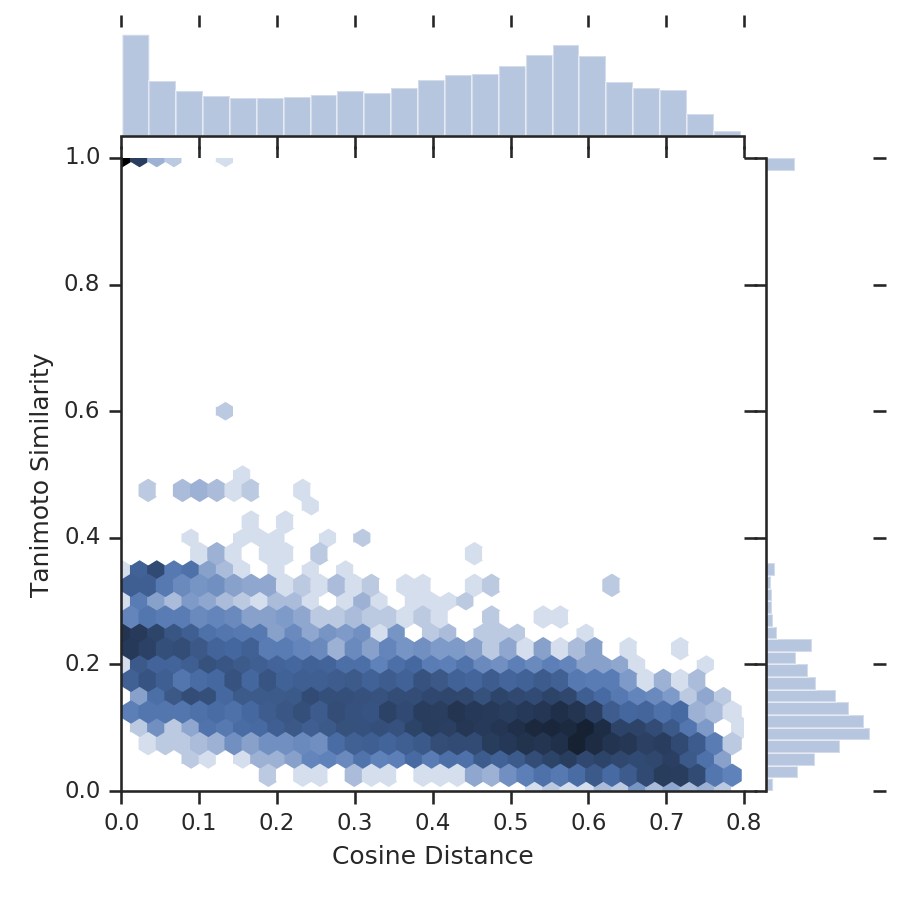}
\caption{\texttt{FC1CC1CC1C2COC21}}
\end{subfigure}
\begin{subfigure}[b]{0.24\textwidth}
\includegraphics[width=\linewidth]{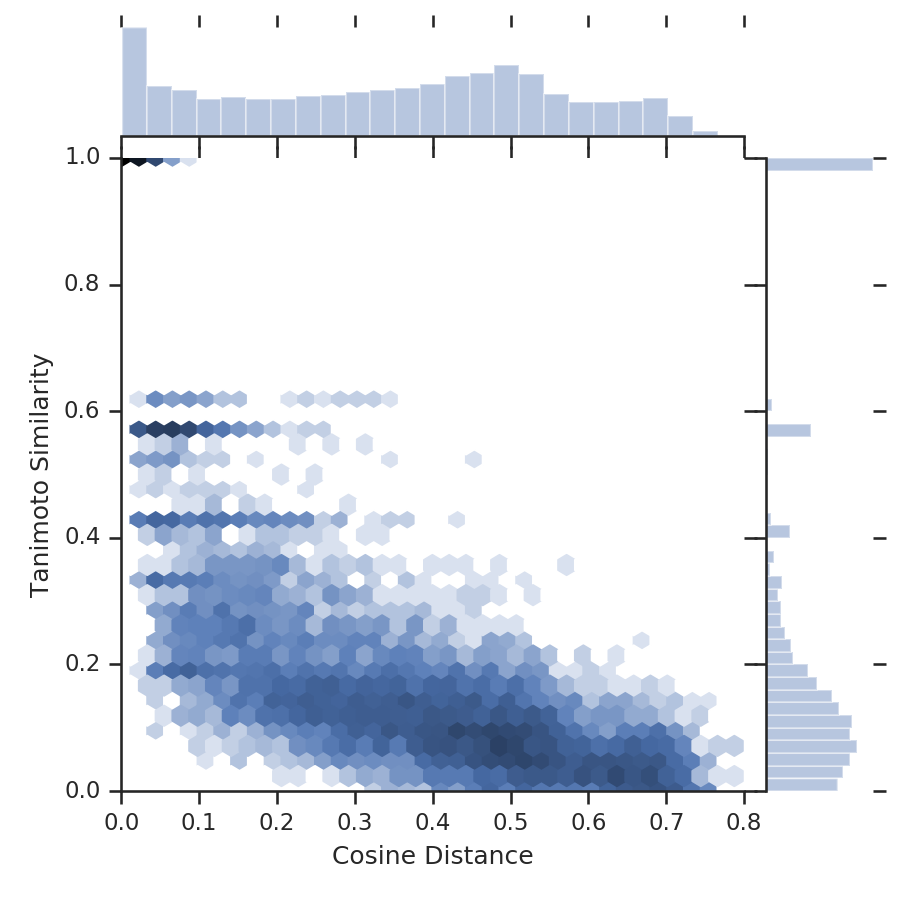}
\caption{\texttt{OC1CC2OC2C1O}}
\end{subfigure}
\begin{subfigure}[b]{0.24\textwidth}
\includegraphics[width=\linewidth]{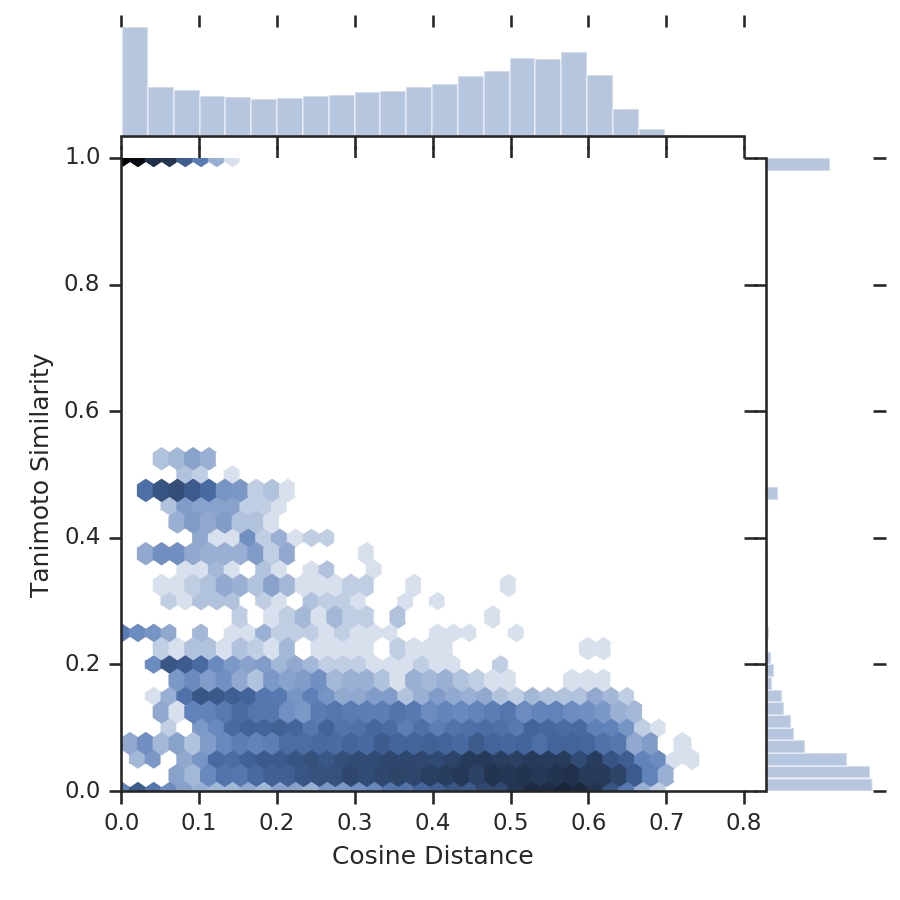}
\caption{\texttt{COC1(C)C(C)C1(C)O}}
\end{subfigure}
\begin{subfigure}[b]{0.24\textwidth}
\includegraphics[width=\linewidth]{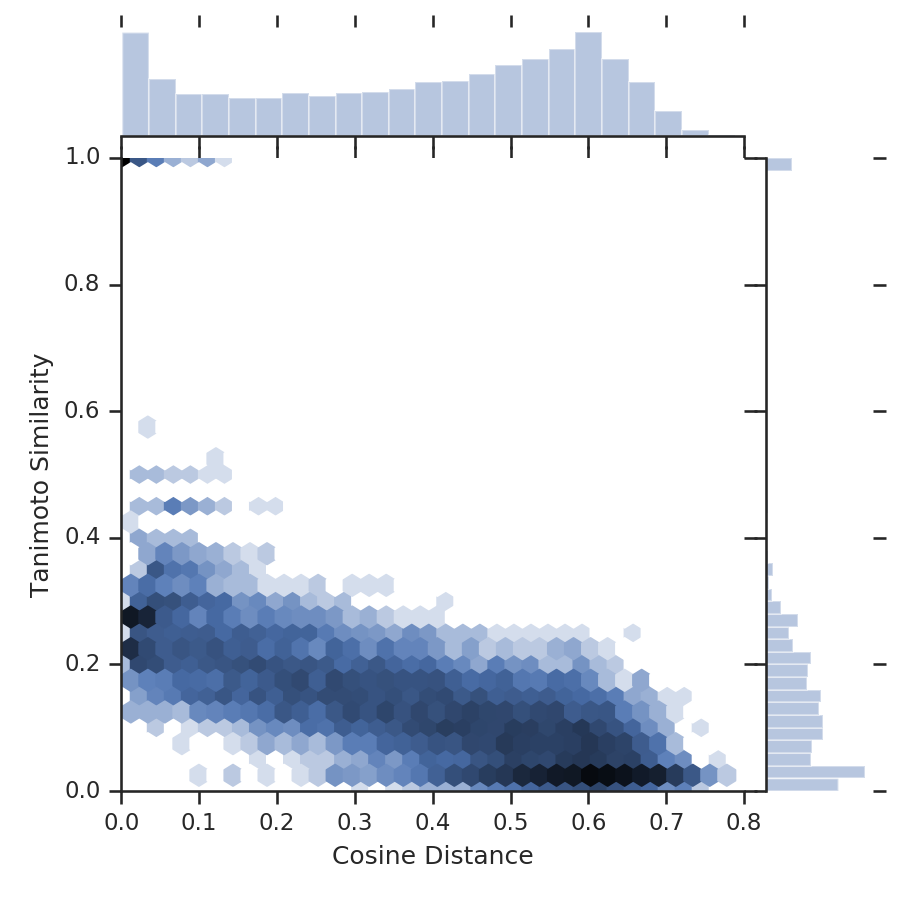}
\caption{\texttt{CC1CC1(O)C1C2CC21}}
\end{subfigure}
\begin{subfigure}[b]{0.24\textwidth}
\includegraphics[width=\linewidth]{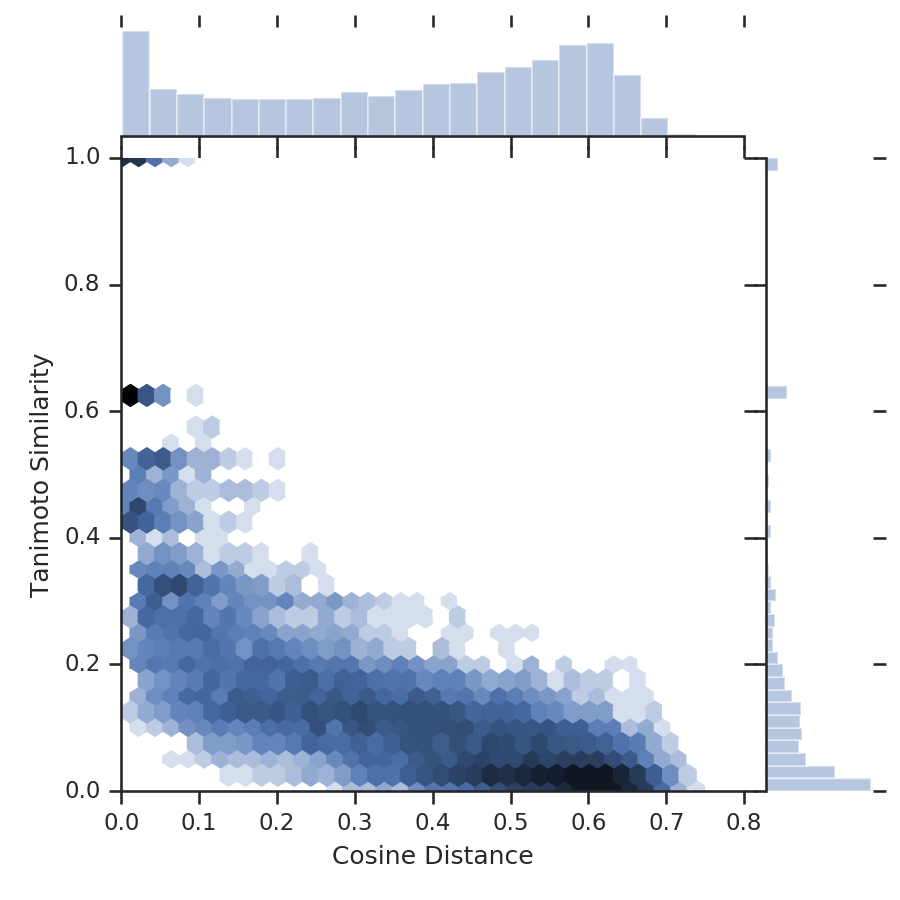}
\caption{\texttt{O=CC\#CC\#CC1CN1}}
\end{subfigure}
\begin{subfigure}[b]{0.24\textwidth}
\includegraphics[width=\linewidth]{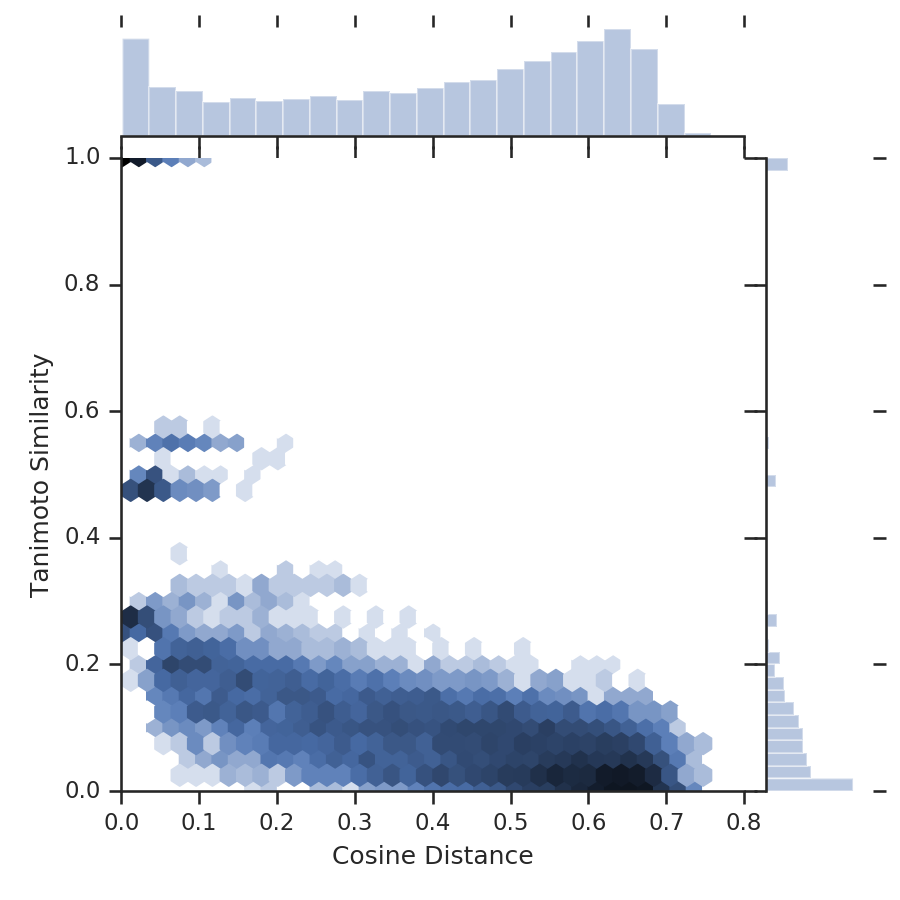}
\caption{\texttt{O=C(CCO)CN1CC1}}
\end{subfigure}
\begin{subfigure}[b]{0.24\textwidth}
\includegraphics[width=\linewidth]{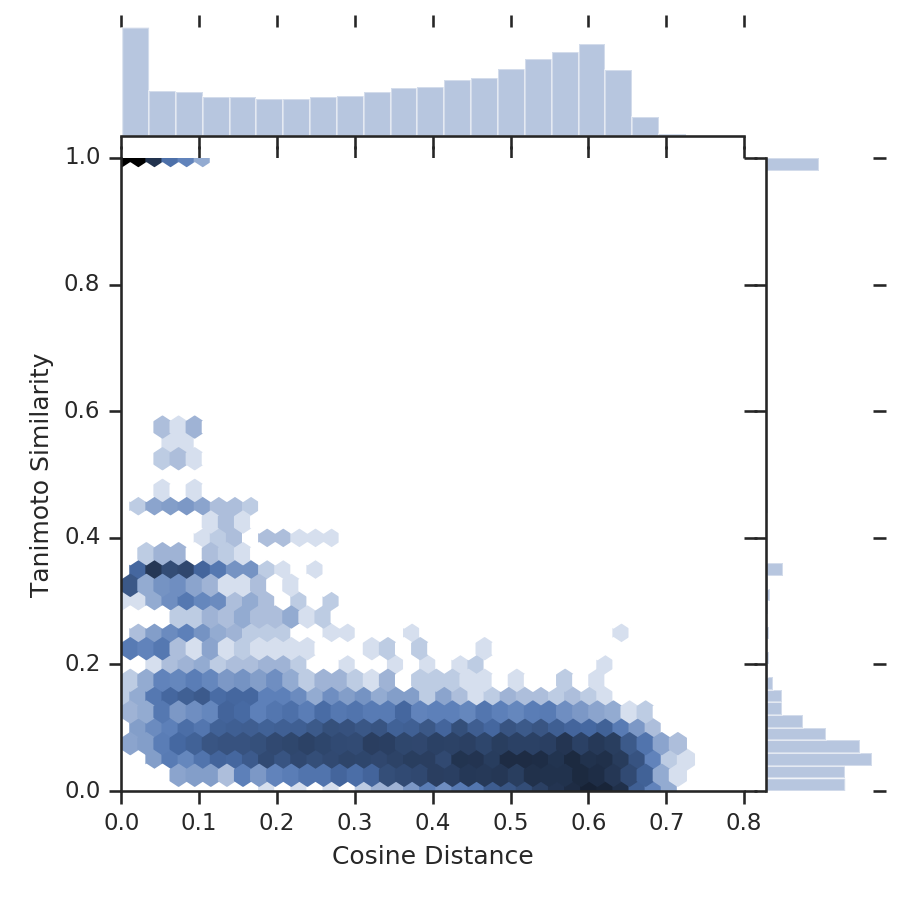}
\caption{\texttt{OC1C2CCC3C(F)C13C2}}
\end{subfigure}
\caption{Tanimoto similarity between molecules decoded from original and perturbed embeddings as a function of cosine distance in the latent space as a function of cosine distance in the latent space. The density color map is logarithmic.}
\label{fig:distance}
\end{center}
\vskip -0.2in
\end{figure*}

\begin{figure*}[tb]
\begin{center}
\begin{subfigure}[b]{0.24\textwidth}
\includegraphics[width=\linewidth]{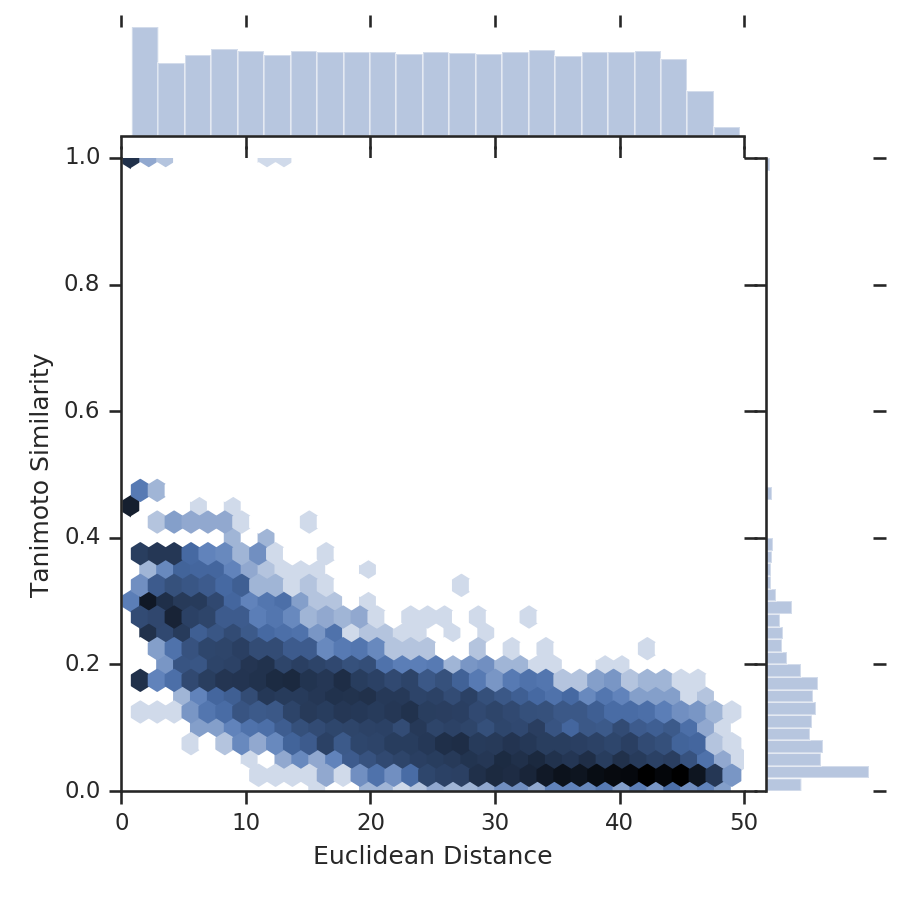}
\caption{\texttt{CCC(C)C1(F)C2OC21C}}
\end{subfigure}
\begin{subfigure}[b]{0.24\textwidth}
\includegraphics[width=\linewidth]{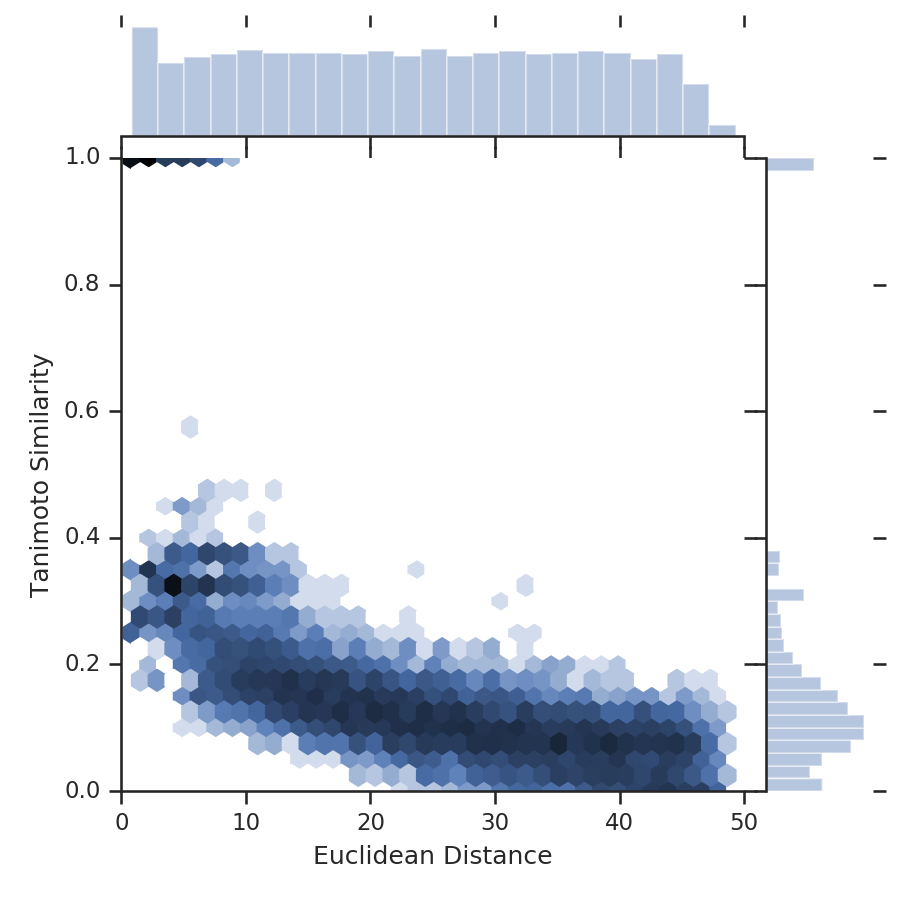}
\caption{\texttt{CC12C(C\#N)C1N1CC12}}
\end{subfigure}
\begin{subfigure}[b]{0.24\textwidth}
\includegraphics[width=\linewidth]{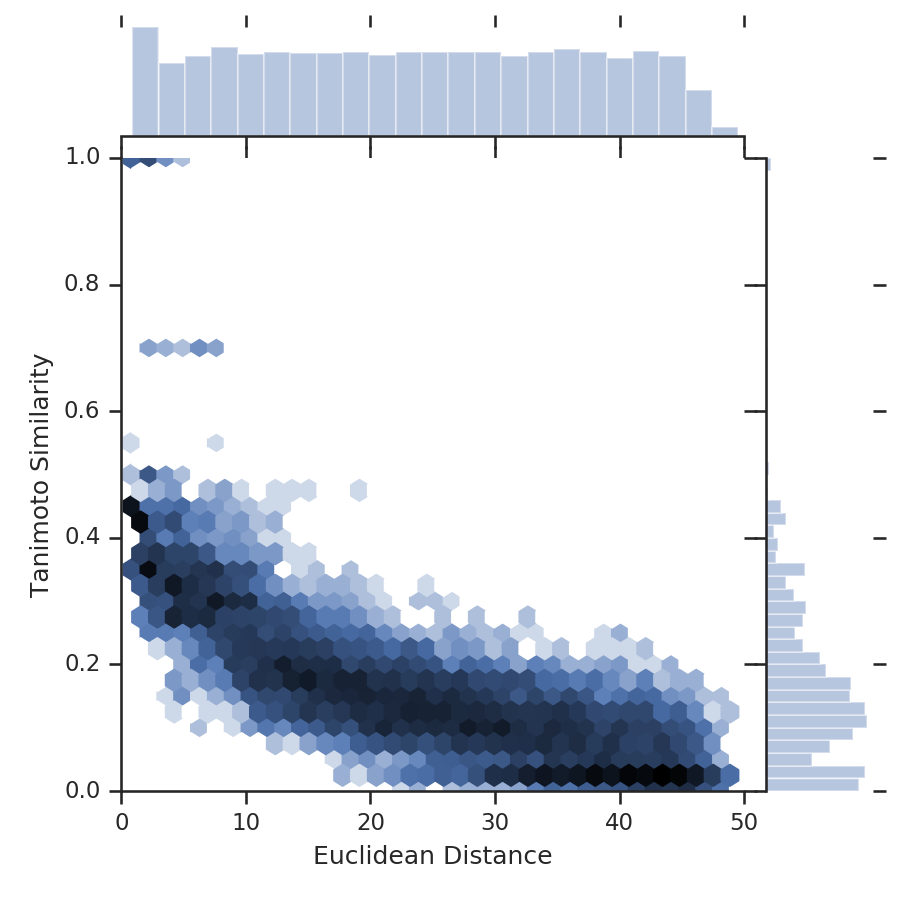}
\caption{\texttt{OCC1CC(O)CO1}}
\end{subfigure}
\begin{subfigure}[b]{0.24\textwidth}
\includegraphics[width=\linewidth]{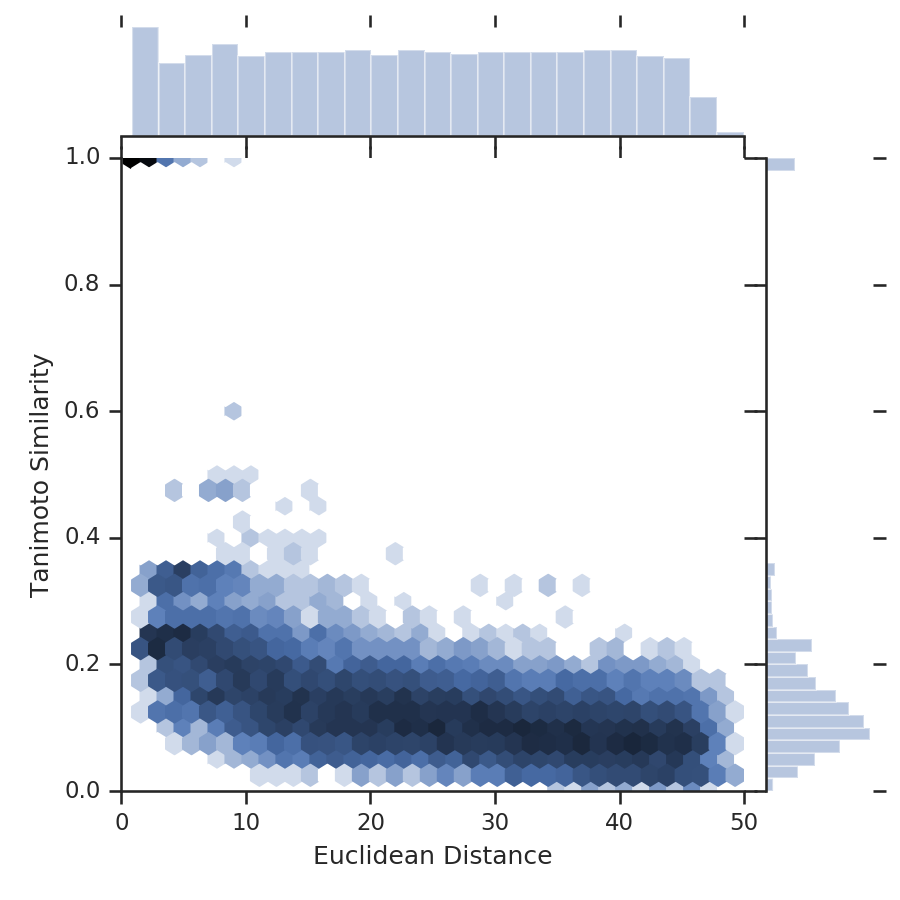}
\caption{\texttt{FC1CC1CC1C2COC21}}
\end{subfigure}
\begin{subfigure}[b]{0.24\textwidth}
\includegraphics[width=\linewidth]{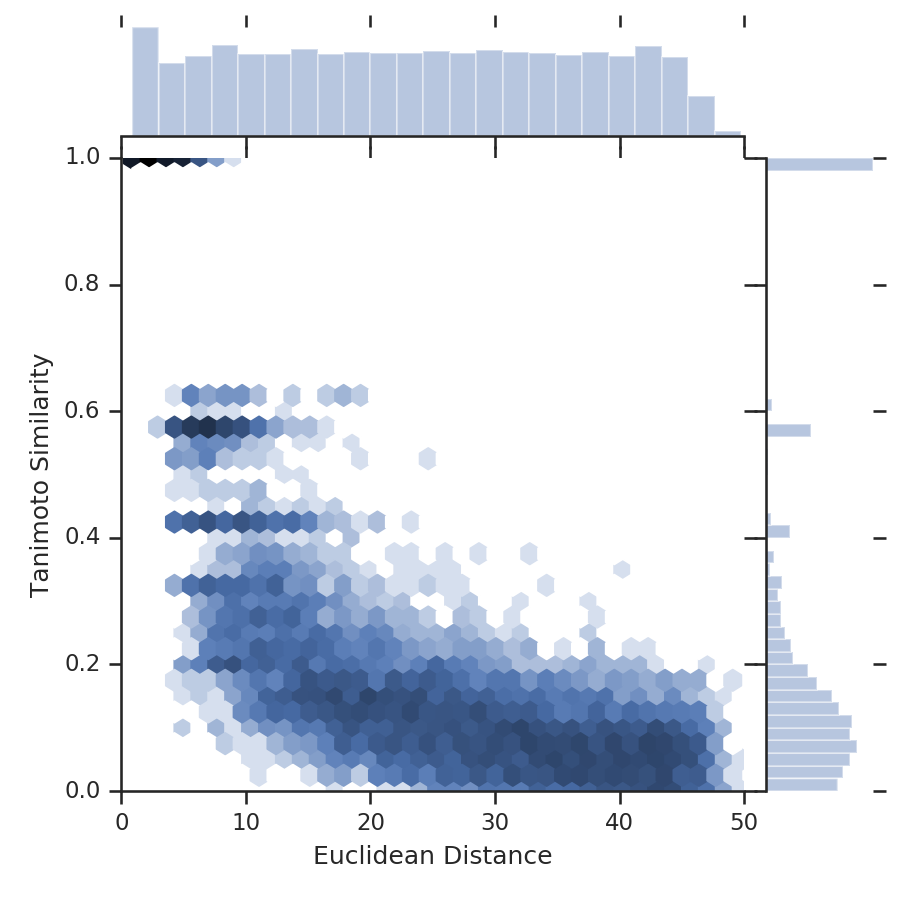}
\caption{\texttt{OC1CC2OC2C1O}}
\end{subfigure}
\begin{subfigure}[b]{0.24\textwidth}
\includegraphics[width=\linewidth]{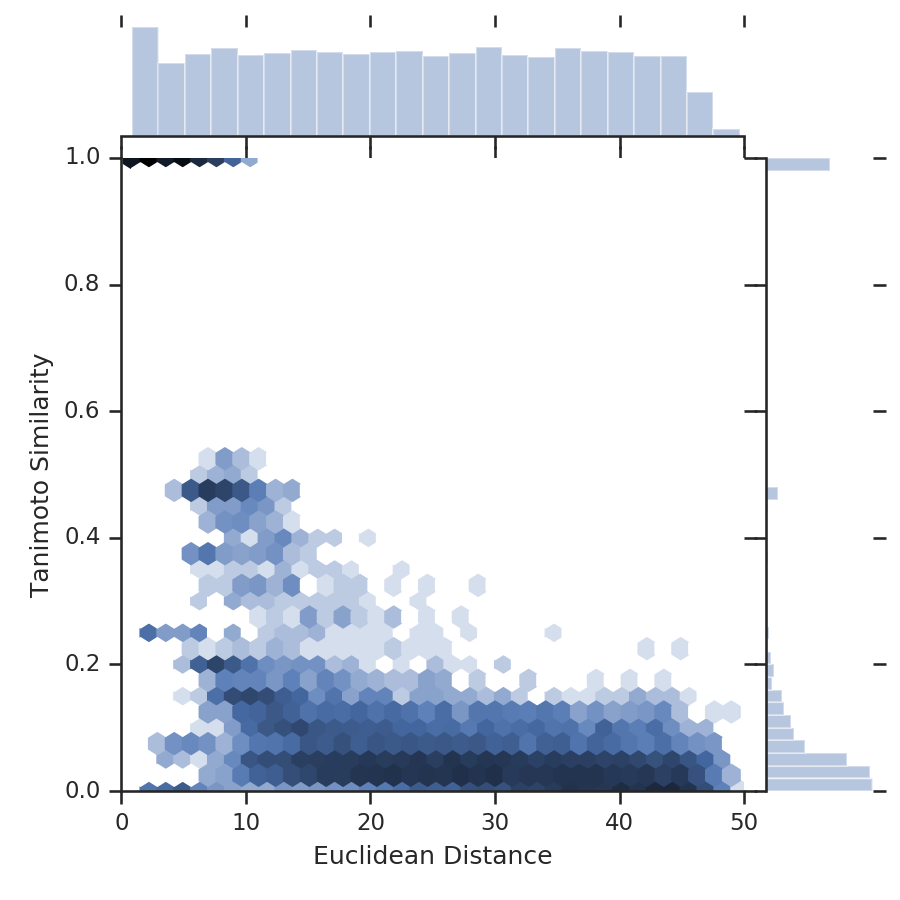}
\caption{\texttt{COC1(C)C(C)C1(C)O}}
\end{subfigure}
\begin{subfigure}[b]{0.24\textwidth}
\includegraphics[width=\linewidth]{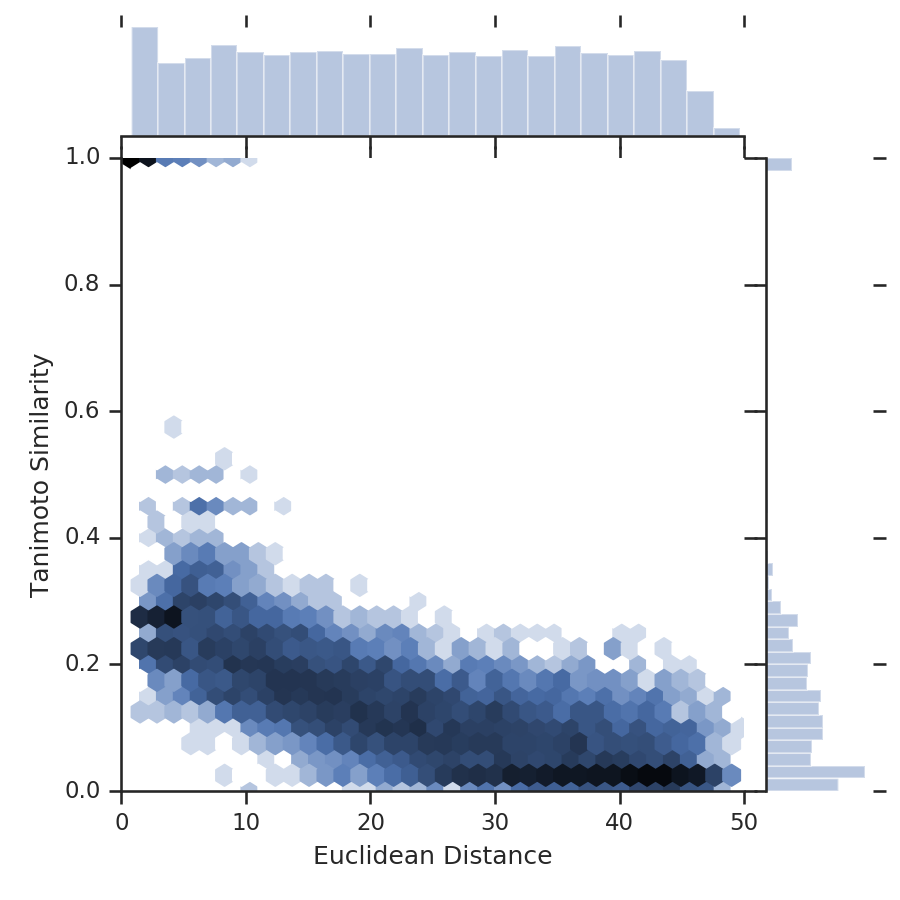}
\caption{\texttt{CC1CC1(O)C1C2CC21}}
\end{subfigure}
\begin{subfigure}[b]{0.24\textwidth}
\includegraphics[width=\linewidth]{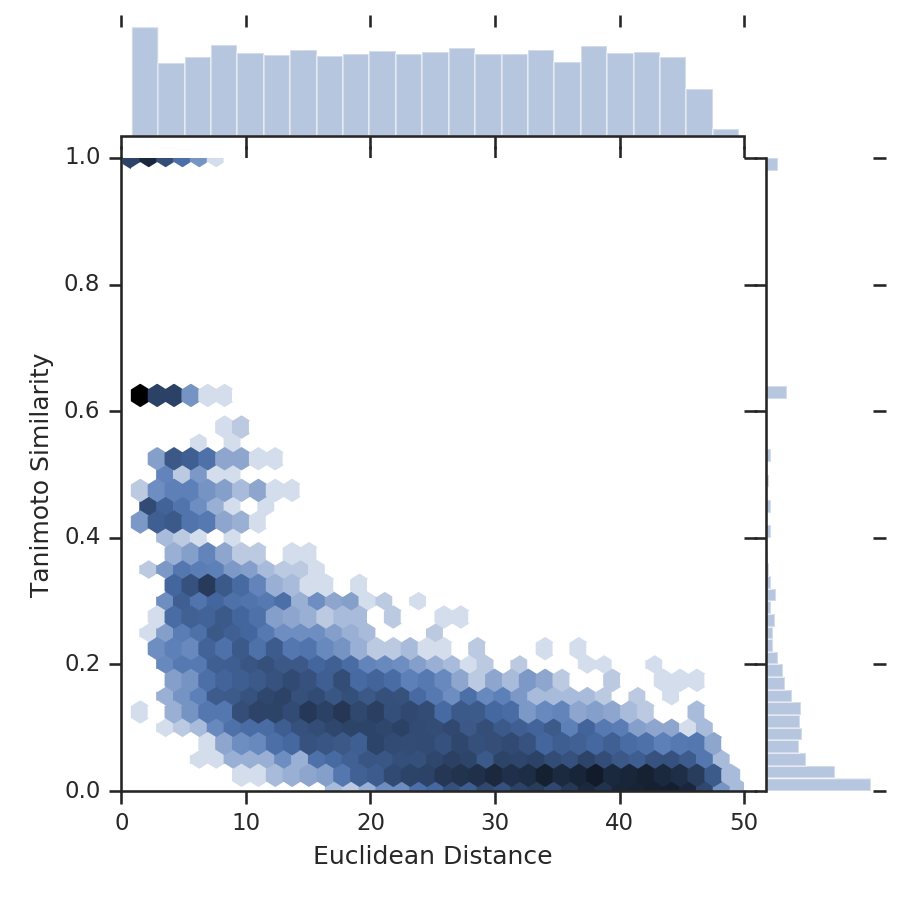}
\caption{\texttt{O=CC\#CC\#CC1CN1}}
\end{subfigure}
\begin{subfigure}[b]{0.24\textwidth}
\includegraphics[width=\linewidth]{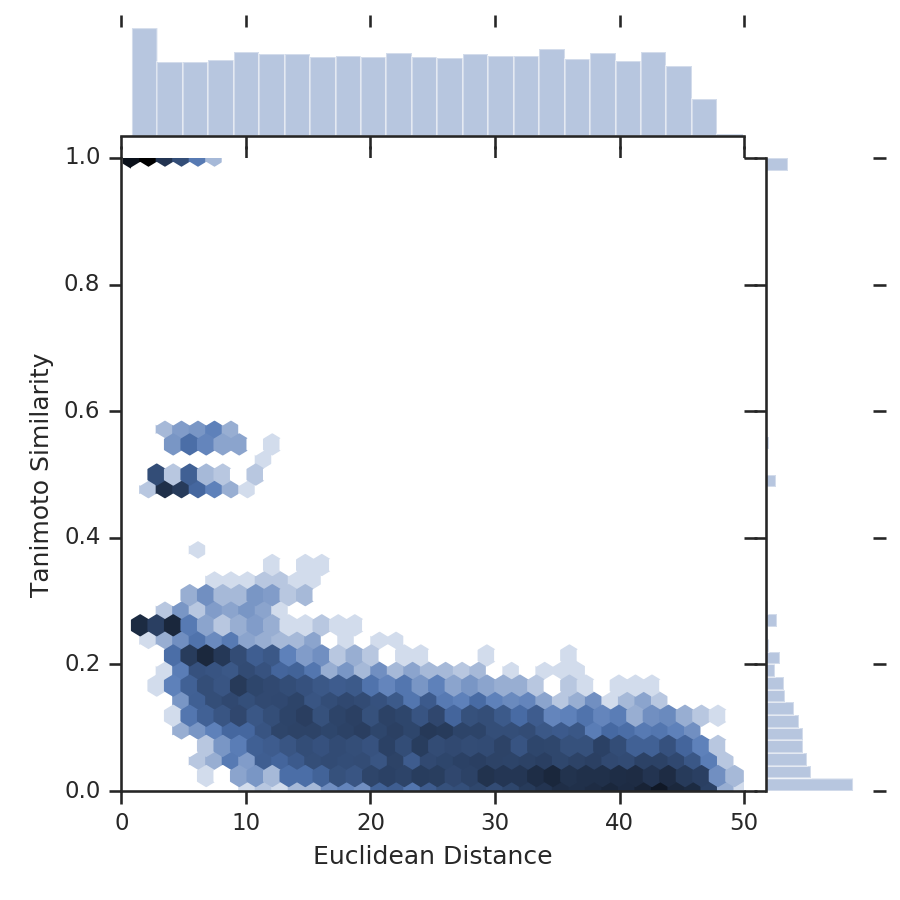}
\caption{\texttt{O=C(CCO)CN1CC1}}
\end{subfigure}
\begin{subfigure}[b]{0.24\textwidth}
\includegraphics[width=\linewidth]{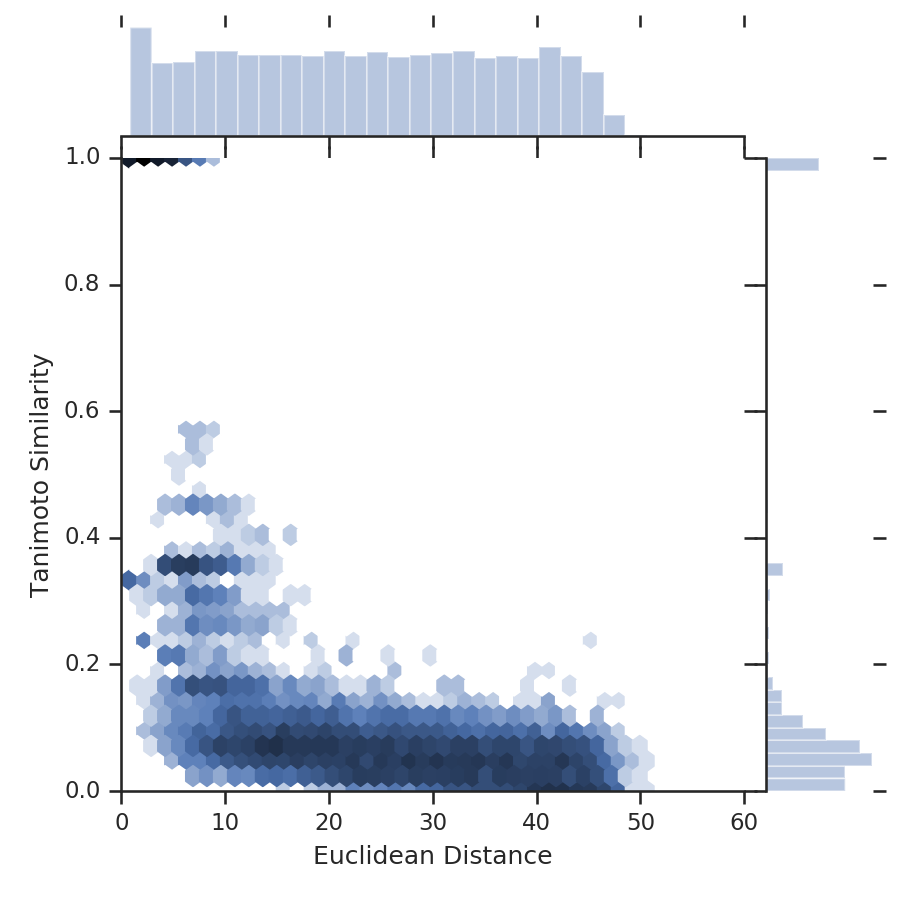}
\caption{\texttt{OC1C2CCC3C(F)C13C2}}
\end{subfigure}
\caption{Tanimoto similarity between molecules decoded from original and perturbed embeddings as a function of cosine distance in the latent space as a function of Euclidean distance in the latent space. The density color map is logarithmic. Similarity was computed on sparse Morgan fingerprints with radius \num{3}.}
\label{fig:distance}
\end{center}
\vskip -0.2in
\end{figure*}

\section{Literature model training and evaluation}

\subsection{JT-VAE}

Code for the JT-VAE model described by \citet{Jin2018-rz} was downloaded from \url{https://github.com/wengong-jin/icml18-jtnn}.
The vocabulary set was extracted from the union of the training, tune and test sets. We used the default settings in the original implementation with our training set to train the model. The model was trained for \num{65000} steps with batch size \num{32}. Note that the data loader in this JT-VAE implementation drops samples if the number of samples in the last batch of each file chunk of data is less than the batch size. Therefore, the number of examples used to compute the reconstruction accuracy on the tune (\num{12736}) and test (\num{12704}) sets are slightly less than the total number of examples in the full tune (\num{13154}) and test (\num{13113}) sets. 

\subsection{GVAE}

Code for the GVAE model described by \citet{Kusner2017-dr} was downloaded from \url{https://github.com/mkusner/grammarVAE}. Training with default parameters was halted by early stopping after 27 epochs. The tune set accuracy associated with the loss used for early stopping was reported to be very low ($<$\num{5}\%), which did not agree with our manual calculation of $\sim$\num{50}\%; we suspect this is because the accuracy calculation used in the code is based on element-wise binarized tensor equality (predicted probabilities are cast to binary predicted labels with a threshold of \num{0.5} by the \texttt{binary\_accuracy} function in Keras), whereas the generated SMILES strings only reflect tensor elements that appear before the first predicted padding element.


\nobibliography{references}
\bibliographystyle{icml2019}